\documentclass[journal]{IEEEtran}
\usepackage{cite}

\ifCLASSINFOpdf
\else
\fi
\usepackage{multirow}

\usepackage[hidelinks]{hyperref}
\usepackage{amssymb}
\usepackage{amsmath}
\usepackage{booktabs}

\usepackage{graphicx}

\begin{document}

\title{ScopeMamba-YOLO: Widening the Perceptual Scope Inward and Outward for Small Object Detection in Remote Sensing Imagery}

\author{Junjie~Fan,
Yijun~Mai,
Linduo~Wei,
Jiayu~Rao,
Junmin~Bao,
Qiushi~Jin,
Guijia~Li,
and~Yong~Qi%
\thanks{This work was supported by the National Key R\&D Program of China under Grant 2019YFE0123800 and the Postgraduate Research \& Practice Innovation Program of Jiangsu Province under Grant 26CXJH1269. \textit{(Corresponding author: Yong Qi.)}}%
\thanks{Junjie Fan, Qiushi Jin, and Yong Qi are with the School of Intellectual Property, Nanjing University of Science and Technology, Nanjing 210094, China (e-mail: junjiefan@njust.edu.cn; 226119010095@njust.edu.cn; qyong@njust.edu.cn).}%
\thanks{Yijun Mai, Jiayu Rao, Junmin Bao, and Guijia Li are with the School of Computer Science and Engineering, Nanjing University of Science and Technology, Nanjing 210094, China (e-mail: maiyijun@njust.edu.cn; jiayurao@njust.edu.cn; sunshine6126@njust.edu.cn; 124106010774@njust.edu.cn).}%
\thanks{Linduo Wei is with the School of Economics and Management, Nanjing University of Science and Technology, Nanjing 210094, China (e-mail: weilinduo@njust.edu.cn).}%
}

\maketitle
\begin{abstract}

Small object detection in unmanned aerial vehicle (UAV) and remote sensing imagery requires
preserving high-resolution detail while modeling long-range context.
Adding a stride-4 detection level and removing the stride-32 stage
benefits tiny targets but weakens peripheral spatial support, whereas
directly inserting selective scanning into the main feature path can
interfere with weak local cues. We propose ScopeMamba-YOLO, built around
an off-path, zero-gated selective-scanning principle that decouples
contextual modeling from the convolutional stream. The principle is
instantiated by a Cascaded Global-Context Module (CGCM) in the backbone
and a Selective-Scan PAN (SS-PAN) in the neck. An Adaptive Multi-scale
Strip (AMS) Block reduces the cost of high-resolution feature extraction,
while a Scale-Adaptive DFL (SA-DFL) head reallocates distributional
support and regression capacity across scales with only 0.008M
additional parameters. Controlled experiments show that matched
main-path selective scanning reduces mAP$_{50}$ by 0.98 pp, whereas
off-path CGCM improves the final configuration by 0.67 pp over the
three-seed no-CGCM mean; operator controls indicate that this gain is
not explained by auxiliary branch capacity alone. ERF analysis further
shows that the complete context pathway increases the peripheral energy
ratio from 0.008 to 0.090 at stride 8. On VisDrone-2019, ScopeMamba-S
achieves 50.8\% mAP$_{50}$ with 3.57M parameters, exceeding YOLOv8s by
10.8 pp while using 32\% of its parameters; ScopeMamba-M reaches
52.6\% mAP$_{50}$ with 6.48M parameters. Consistent improvements are
also observed on AI-TOD, especially for very-tiny and tiny objects.
\end{abstract}

\begin{IEEEkeywords}
Small object detection, UAV imagery, remote sensing, state space model, Mamba, receptive field, YOLO.
\end{IEEEkeywords}

%
\IEEEpeerreviewmaketitle

\section{Introduction}

\begin{figure}[htbp]
\centering
\includegraphics[width=\linewidth]{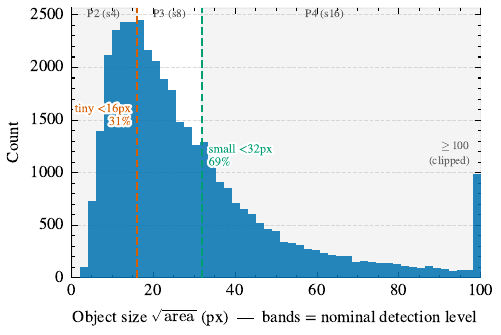}
\caption{Bounding box size distribution (square root of area) in the VisDrone-2019 dataset. The vertical dashed lines delineate the nominal detection level bands, matching specific object sizes to their nominally associated detection strides (e.g., P2 for targets $<16$ pixels, P3 for targets $<32$ pixels).}
\label{fig:bbox_area_hist}
\end{figure}

\IEEEPARstart{O}{bject} detection in remote sensing imagery is a fundamental prerequisite for a broad spectrum of Earth-observation applications, including urban planning, traffic monitoring, disaster assessment, precision agriculture, and maritime surveillance~\cite{cheng2016survey,li2020object,xia2018dota}. Benefiting from deep convolutional neural networks (CNNs), generic detectors---whether two-stage pipelines built upon region proposals~\cite{ren2017faster,he2017mask} or one-stage frameworks that regress objects directly from dense feature maps~\cite{liu2016ssd,lin2017focal,tian2019fcos,redmon2016you}---have achieved remarkable success on natural-image benchmarks such as MS COCO~\cite{lin2014microsoft}. When transplanted to aerial scenarios, however, their performance degrades substantially, and the gap is most pronounced for small objects. In images captured from high altitudes with wide fields of view, vehicles, ships, and pedestrians frequently occupy fewer than $32\times32$ pixels, and often fewer than $16\times16$~\cite{cheng2023towards,wang2021tiny}. On UAV and tiny-object benchmarks such as VisDrone~\cite{zhu2022detection} and AI-TOD~\cite{wang2021tiny}, such instances constitute the overwhelming majority of annotations (Fig.~\ref{fig:bbox_area_hist}), making small-object detection a central challenge for practical remote sensing systems.

The difficulty stems from an inherent contradiction between what a small object can offer and what recognizing it demands. On the one hand, a small object carries extremely limited appearance evidence---sparse texture, ambiguous boundaries, and a low signal-to-noise ratio---so its discriminative features are easily attenuated or completely submerged during the repeated downsampling of hierarchical backbones~\cite{lin2017feature,cheng2023towards,sun2026weyolo}. On the other hand, correctly recognizing such a weak signal requires \emph{more}, rather than less, supporting information: fine-grained local detail is needed to distinguish a vehicle from a rooftop fixture, while long-range contextual cues (roads, harbors, parking lots) are needed to suppress the background clutter that dominates aerial scenes and occupies the vast majority of image area~\cite{li2020object,zhang2024ffca,liu2025esod}. In other words, the \emph{perceptual scope} of the detector must be widened in two opposite directions simultaneously: \emph{inward}, toward higher-resolution, fine-grained representations of the object itself; and \emph{outward}, toward scene-level contextual dependencies that reach far beyond it.

Existing solutions often emphasize one side of this resolution--context
trade-off. To preserve fine-grained information, prior works exploit
multi-scale feature pyramids and their bidirectional variants
~\cite{lin2017feature,liu2018path,tan2020efficientdet}, slicing-based
high-resolution inference~\cite{akyon2022slicing}, and small-object-friendly
label assignment~\cite{xu2022detecting,xu2022rfla}. For
lightweight aerial detectors, a common structural choice is to add a
stride-4 (P2) detection level while removing the stride-32 (P5) stage
~\cite{zhu2021tph,hua2026sfbf,zheng2025smnyolo}. This reallocation
substantially improves small-object accuracy in our ablations, but it
also removes the deepest feature stage and reduces peripheral spatial
support, as quantified later by the ERF analysis. Complementary
approaches enlarge contextual support through coarse-to-fine processing
~\cite{yang2019clustered,li2020density,huang2022ufpmp}, attention
mechanisms~\cite{hu2018squeeze,woo2018cbam}, transformer prediction
heads~\cite{zhu2021tph}, and large-kernel or selective receptive-field
designs~\cite{li2023large,cai2024poly}. Self-attention
~\cite{vaswani2017attention} provides global modeling but incurs
quadratic complexity with respect to token number, which becomes costly
on the high-resolution feature maps required for small-object detection.
The resulting design challenge is therefore to preserve high-resolution
detail while retaining broader contextual support at manageable cost.

Recently, state space models (SSMs)~\cite{gu2022efficiently}, and in particular the selective SSM Mamba~\cite{gu2023mamba,dao2024transformers}, have emerged as a compelling alternative for modeling long-range dependencies with linear complexity. Vision adaptations such as Vision Mamba~\cite{zhu2024vision} and VMamba~\cite{liu2024vmamba} extend this capability to dense visual representations, which has driven a rapid migration of Mamba into remote sensing~\cite{chen2024rsmamba,zhao2024rs,chen2024changemamba} and, more recently, into YOLO-style small object detectors~\cite{wang2025mamba,wu2025mv,wang2026hedge,hua2026sfbf,rong2025ssmnet,feng2026mvmamba,jiang2026lghvss}. In principle, such a linear-cost global operator is well suited to recovering the broader contextual support weakened by high-resolution scale reallocation. In practice, however, existing designs typically embed selective scanning directly into the main feature path of the backbone or neck. Considerable effort has therefore been devoted to refining the scan order and block design~\cite{shi2024msvmamba,xie2024quadmamba}, while the question of \emph{where} the scan should sit relative to the detail-carrying stream has received much less attention. On a shared pathway, the scan competes with fine-grained convolutions for limited channel capacity and propagates state over feature maps in which background tokens vastly outnumber the few tokens belonging to a small object. For tiny targets, this coupling creates a risk that long-range aggregation may dilute weak local evidence rather than reinforce it.

Our controlled experiments provide empirical evidence consistent with this concern. In an earlier matched configuration, direct main-path selective scanning reduces mAP$_{50}$ by 0.98 pp at unchanged computation, whereas under the final configuration the off-path CGCM improves mAP$_{50}$ by 0.67 pp over the no-CGCM control. Although these observations come from different stages of the design trajectory, together they motivate a core design principle: decoupling long-range context modeling from the detail-carrying main path, allowing selective scanning to complement rather than replace the local representation stream.

To resolve this contradiction, we propose ScopeMamba-YOLO, an efficient
detector for UAV and remote sensing imagery built on a single
through-line: \emph{reallocate} the detection scales toward the
resolution regime where small objects survive, \emph{read} those
high-resolution features anisotropically at reduced cost, \emph{reach}
back out to broader scene context through off-path selective scanning,
and \emph{regress} boxes with a distributional regression budget adapted
to each detection scale. Rather than replacing main-path convolutions
with state space blocks, we introduce an \emph{off-path, zero-gated
selective-scanning} principle: the scan runs on a low-cost side
branch---pooled to $\tfrac14$ resolution in the backbone and narrowed in
the neck---and re-enters the main path only through zero-initialized
gates, so that the network is functionally identical to its convolutional
skeleton at initialization and the convolutional stream retains full
ownership of the representation. The principle is instantiated by the
Cascaded Global-Context Module (CGCM) in the backbone and the
Selective-Scan PAN (SS-PAN) in the neck, which broaden the long-range
spatial support weakened by removing the P5 stage while retaining the
fine-grained convolutional stream. The Adaptive Multi-scale Strip (AMS)
Block reduces the resource cost of the reallocated backbone through
anisotropic feature extraction, and the Scale-Adaptive DFL (SA-DFL)
head assigns scale-specific distributional support and regression
capacity with only 0.008M additional parameters. Four variants
(N, S, M, and L) span a wide accuracy--efficiency range. On
VisDrone-2019, ScopeMamba-N exceeds YOLOv8n by 9.8 mAP$_{50}$ points
with 34\% of its parameters, ScopeMamba-S exceeds YOLOv8s by 10.8
points with 32\% of its parameters, and ScopeMamba-M reaches the
accuracy level of the recent Mamba-based HEdge-MamYOLO~\cite{wang2026hedge}
with less than one-third of its parameters (6.48M versus 20.8M);
consistent gains are also observed on AI-TOD.

The main contributions of this article are summarized as follows.
\begin{enumerate}

    \item We propose ScopeMamba-YOLO, an efficient detector tailored for
    small objects in UAV and remote sensing imagery. By decoupling
    scene-level context modeling from the fine-grained convolutional stream,
    the proposed framework jointly preserves high-resolution object evidence
    and broadens contextual support, achieving a favorable
    accuracy--parameter trade-off across four scalable variants
    (N, S, M, and L).

    \item We formulate the off-path, zero-gated selective-scanning principle, instantiated by the CGCM in the backbone and the SS-PAN in the neck. It broadens long-range spatial support without directly replacing the fine-grained convolutional stream. Controlled in-path and off-path experiments, operator controls, and effective-receptive-field analysis provide complementary evidence for this integration strategy and indicate that the observed off-path gain is not explained by auxiliary branch capacity alone.

    \item We design the AMS Block as a resource-aware feature extraction engine. Its adaptive strip kernels and cross-directional gating capture anisotropic structures at close to depthwise-line cost, reducing the parameters and GFLOPs of the reallocated baseline by 24\% and 9\%, respectively, while maintaining comparable accuracy and providing part of the resource budget for the context pathways.
    
    \item We introduce the SA-DFL head, which assigns each detection scale
    its own distributional regression budget. Because the per-scale bin
    allocation also changes the regression-branch width, SA-DFL jointly
    reallocates distributional support and regression capacity across
    feature levels. Among the proposed design changes, SA-DFL provides the
    largest marginal mAP$_{50}$ gain per added parameter, while additional
    regression and quality-branch controls favor this scale-adaptive
    regression design over the tested alternatives under our experimental
    setting.
\end{enumerate}

\section{Related Work}

\subsection{Small Object Detection in Aerial Imagery}

Object detection has matured along two-stage~\cite{ren2017faster,he2017mask} and one-stage~\cite{liu2016ssd,lin2017focal,tian2019fcos} paradigms, yet both were designed for natural images and degrade sharply on the tiny, densely distributed instances that dominate aerial benchmarks such as VisDrone~\cite{zhu2022detection} and AI-TOD~\cite{wang2021tiny}. Research on aerial small objects has followed several complementary lines. One raises the effective input resolution through slicing-aided inference~\cite{akyon2022slicing}; a second reforms the supervision with scale-robust label assignment~\cite{xu2022detecting,xu2022rfla}; a third exploits the sparsity of foreground in aerial scenes by routing computation to object-dense subregions in a coarse-to-fine manner~\cite{yang2019clustered,li2020density,huang2022ufpmp,yang2022querydet,liu2025esod}. For UAV deployment, where onboard compute and real-time inference impose strict limits, a fourth line redesigns the detector itself to be small and fast~\cite{zhang2024ffca,zheng2025smnyolo,hou2025mfelyolo,sun2026weyolo}.

Within this last line, a widely adopted structural prior is to reallocate the detection scales toward higher resolution. Since a small object survives only a few downsampling stages before its evidence vanishes, the shallow stride-4 P2 level is attached to recover fine spatial detail, while the deepest stride-32 P5 level---which contributes little to sub-16-pixel targets---is pruned to bound the added cost~\cite{zhu2021tph,zhang2024ffca,hua2026sfbf,zheng2025smnyolo}. We adopt the same reallocation prior and, through the AMS Block, further trim the parameter overhead it entails. The prior, however, is not free in a way that is rarely made explicit: abruptly discarding the deepest stage shrinks the overall receptive field of the network, so the global context that P5 used to provide is lost precisely when it is most needed to separate small objects from clutter. The following two directions attempt to remedy this deficiency.

\subsection{Context Modeling and Receptive Field Expansion}

To compensate for a limited or deliberately reduced receptive field, a large body of work enlarges the spatial context accessible to each detection feature. Along the convolutional line, large-kernel and selective-kernel designs such as LSKNet~\cite{li2023large} and PKINet~\cite{cai2024poly} widen the effective receptive field for remote sensing objects, while channel and spatial attention modules~\cite{hu2018squeeze,woo2018cbam} and transformer prediction heads~\cite{zhu2021tph} inject scene-level cues. Along the fusion line, feature pyramids and their bidirectional variants~\cite{lin2017feature,liu2018path,tan2020efficientdet}, together with context-aware necks such as FFCA-YOLO~\cite{zhang2024ffca} and frequency-aware fusion designs~\cite{wang2026hedge}, recombine multi-level features to propagate semantics across scales. These strategies are effective but share two structural limits. First, feature fusion redistributes information across pyramid levels but does not by itself guarantee recovery of the long-range spatial dependencies weakened by deep-stage pruning. Second, self-attention~\cite{vaswani2017attention,dosovitskiy2021image,liu2021swin} and DETR-style detectors~\cite{carion2020end,zhu2021deformable,zhao2024detrs} can model genuinely global relations, but their quadratic cost is prohibitive on the high-resolution feature maps that small object detection requires. This motivates global-context mechanisms whose computational cost remains manageable on the high-resolution feature maps required for tiny-object detection.

\subsection{State Space Models for Visual Representation}

State space models offer exactly such a mechanism. The structured SSM S4~\cite{gu2022efficiently} and the selective, hardware-aware Mamba~\cite{gu2023mamba,dao2024transformers} capture long-range dependencies with linear complexity, and vision adaptations such as Vision Mamba~\cite{zhu2024vision} and VMamba~\cite{liu2024vmamba} extend this global receptive field to dense prediction, with subsequent work refining the serialization of 2-D features into scan sequences~\cite{shi2024msvmamba,xie2024quadmamba}. Their efficiency has driven wide adoption in remote sensing for classification~\cite{chen2024rsmamba}, dense prediction~\cite{zhao2024rs}, change detection~\cite{chen2024changemamba}, and pan-sharpening~\cite{he2025pan}, as well as a fast-growing family of Mamba-based detectors: Mamba YOLO~\cite{wang2025mamba} couples selective scanning with a real-time head, MiM-ISTD~\cite{chen2024mimistd} adapts it to infrared small targets, and SODMAMBA-DETR~\cite{sun2025sodmamba}, SSMNet~\cite{rong2025ssmnet}, MV-YOLO~\cite{wu2025mv}, MVMamba~\cite{feng2026mvmamba}, LGHVSS-Mamba-YOLO~\cite{jiang2026lghvss}, PGI-ViMamba~\cite{liu2025pgivimamba}, HEdge-MamYOLO~\cite{wang2026hedge}, and LEM-YOLO~\cite{yao2025lemyolo} tailor SSM blocks to small targets in aerial and remote sensing imagery, from UAV scenes to ship detection.

These detectors confirm the value of linear-complexity long-range
modeling, while much of their design effort focuses on the
operator---which directions to scan, how to combine selective scanning
with convolution or attention inside a block, and how to reduce its
cost. In many existing designs, the SSM replaces or augments blocks on
the main feature path of the backbone or neck, so sparse small-object
tokens share representation capacity with local feature extraction and
participate in state propagation over feature maps in which background
tokens are numerically dominant. For tiny targets, this coupling creates
a risk that weak local evidence may be diluted during long-range
aggregation. Relatively few designs explicitly decouple the global scan
from the detail-carrying stream, and controlled studies that isolate
this placement dimension remain limited. Motivated by this gap, we move
selective scanning off the main path and use CGCM and SS-PAN to broaden
the long-range spatial support weakened by P5 removal while retaining
the high-resolution feature stream. We then evaluate the consequences
of this integration strategy through controlled experiments.

\section{Proposed Method}
\label{sec:method}

\begin{figure*}[!t]
\centering
\includegraphics[width=\textwidth]{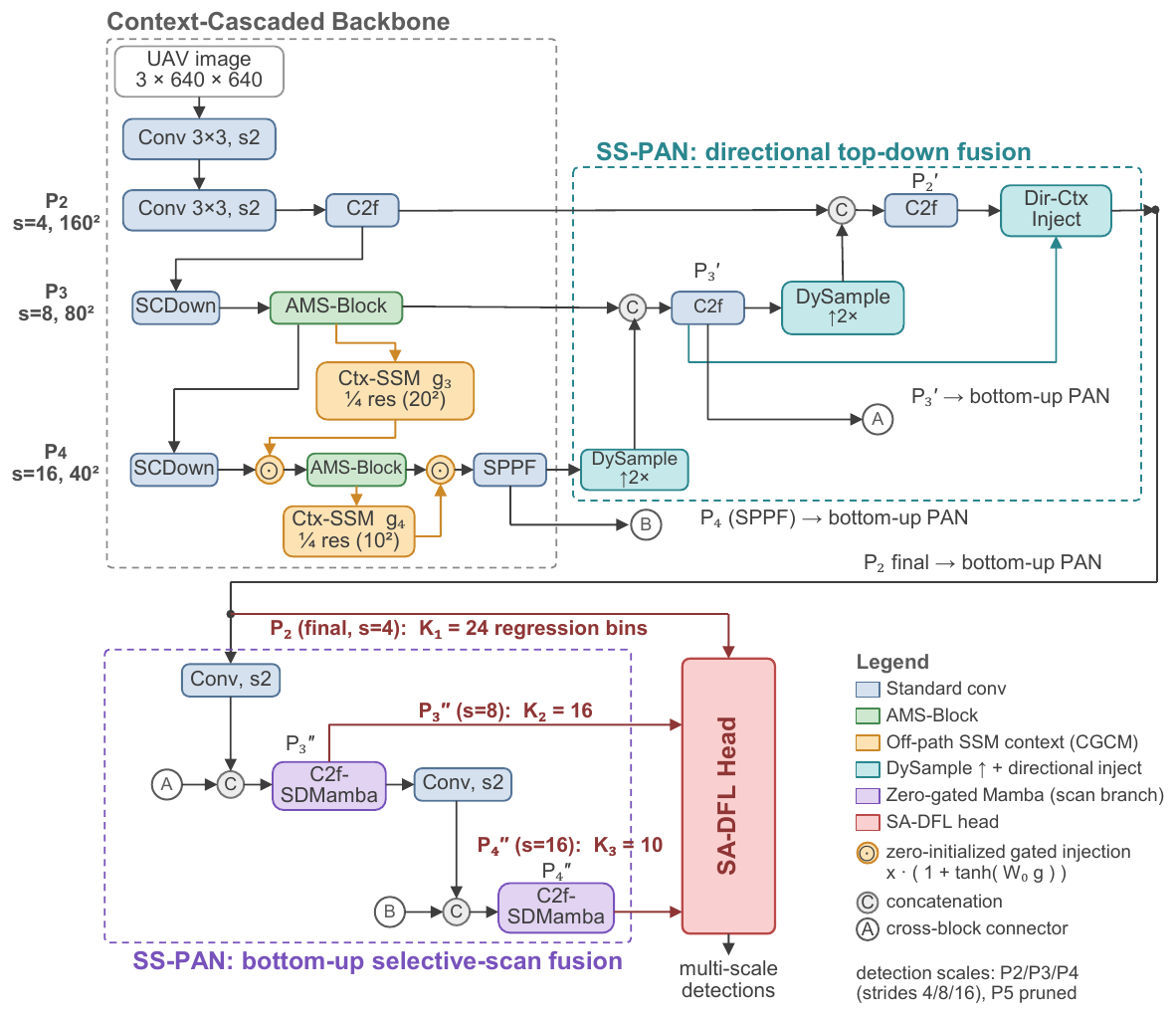}
\caption{Overall architecture of ScopeMamba-YOLO. The network detects on
P2/P3/P4 (strides 4/8/16) with the P5 stage removed (high-resolution
reallocation). In the Context-Cascaded Backbone, AMS-Blocks perform
anisotropic multi-scale reading, while two off-path context streams
(CGCM context streams $g_3$, $g_4$; orange) run selective scans on
$\tfrac14$-resolution features and modulate the main path through
zero-initialized gates ($\odot$). The SS-PAN neck consists of a top-down pathway using DySample
up-sampling and directional context injection (cyan), together with a
bottom-up pathway whose P3/P4 fusion nodes carry zero-gated
selective-scan branches (purple). The SA-DFL head performs scale-adaptive
distributional box regression with per-scale bin budgets
$K{=}[24,16,10]$.}
\label{fig:overall}
\end{figure*}

\subsection{Overview and Scale Reallocation}
\label{sec:method_overview}

ScopeMamba-YOLO is built on the YOLOv8 skeleton and follows a single
through-line: \emph{Reallocate} the detection scales toward the resolution
regime where small objects survive, \emph{Read} those high-resolution
features with anisotropy-aware operators at reduced cost, \emph{Reach}
back out to scene-level context through off-path selective scanning, and
\emph{Regress} boxes with a distributional regression budget matched to each scale.
Fig.~\ref{fig:overall} shows the resulting architecture; this subsection
fixes the scale layout that everything else is built on.

\subsubsection{High-Resolution Reallocation (HRR)}
Following the established practice for aerial small-object detection
\cite{zhu2021tph,hua2026sfbf}, we add a stride-4 (P2, $160\times160$ at a
$640\times640$ input) detection branch and remove the stride-32 (P5)
stage and its detection head entirely, detecting on P2/P3/P4 with strides
$\{4,8,16\}$. We treat this reallocation as an adopted prior rather than a
contribution: the bounding-box statistics of VisDrone
(Fig.~\ref{fig:bbox_area_hist}) show that the overwhelming majority of
instances fall below the area regime for which a stride-32 grid retains
even a single-cell response, whereas targets larger than the P4 regime
are rare. 

HRR therefore provides a suitable base allocation of spatial budget for
the present small-object setting, but it is not free. Pruning the P5
pathway removes roughly $7$M parameters from the YOLOv8s skeleton and
also weakens the peripheral spatial support available to the remaining
high-resolution detection features, as quantified later by the ERF
analysis. The detector therefore gains resolution at the cost of reduced
contextual reach. Sections~\ref{sec:method_offpath} and
~\ref{sec:method_neck} address this trade-off by broadening long-range
spatial support without reverting the high-resolution scale allocation;
Section~\ref{sec:method_ams} first reduces the resource cost of that
allocation.

\subsubsection{Data flow}
The backbone (layers 0--11 in Fig.~\ref{fig:overall}) is a
\emph{Context-Cascaded Backbone}: a convolutional spatial main path
(stem $\rightarrow$ C2f at P2 $\rightarrow$ SCDown $\rightarrow$
AMS-Block at P3 $\rightarrow$ SCDown $\rightarrow$ AMS-Block
$\rightarrow$ SPPF at P4), flanked by two low-resolution context streams
$g_3$ and $g_4$ that re-enter the main path only through
zero-initialized multiplicative gates. Because the network no longer has
a stride-32 stage, SPPF is relocated to the end of the P4 stage. The neck
first runs a top-down pathway that fuses P4$\rightarrow$P3
$\rightarrow$P2 using DySample up-sampling and enriches the P2 map with
directional semantics from the P3 level, followed by the bottom-up
pathway of SS-PAN, whose P3 and P4 fusion nodes carry zero-gated
selective-scan branches. The SA-DFL head consumes the three resulting maps
($P_2^{\mathrm{final}}$, $P_3''$, $P_4''$), which at a $640^2$ input
yield $160^2{+}80^2{+}40^2 = 33{,}600$ anchors.

\subsection{AMS-Block: Anisotropic Multi-Scale Reading}
\label{sec:method_ams}

\begin{figure}[!t]
\centering
\includegraphics[width=\columnwidth]{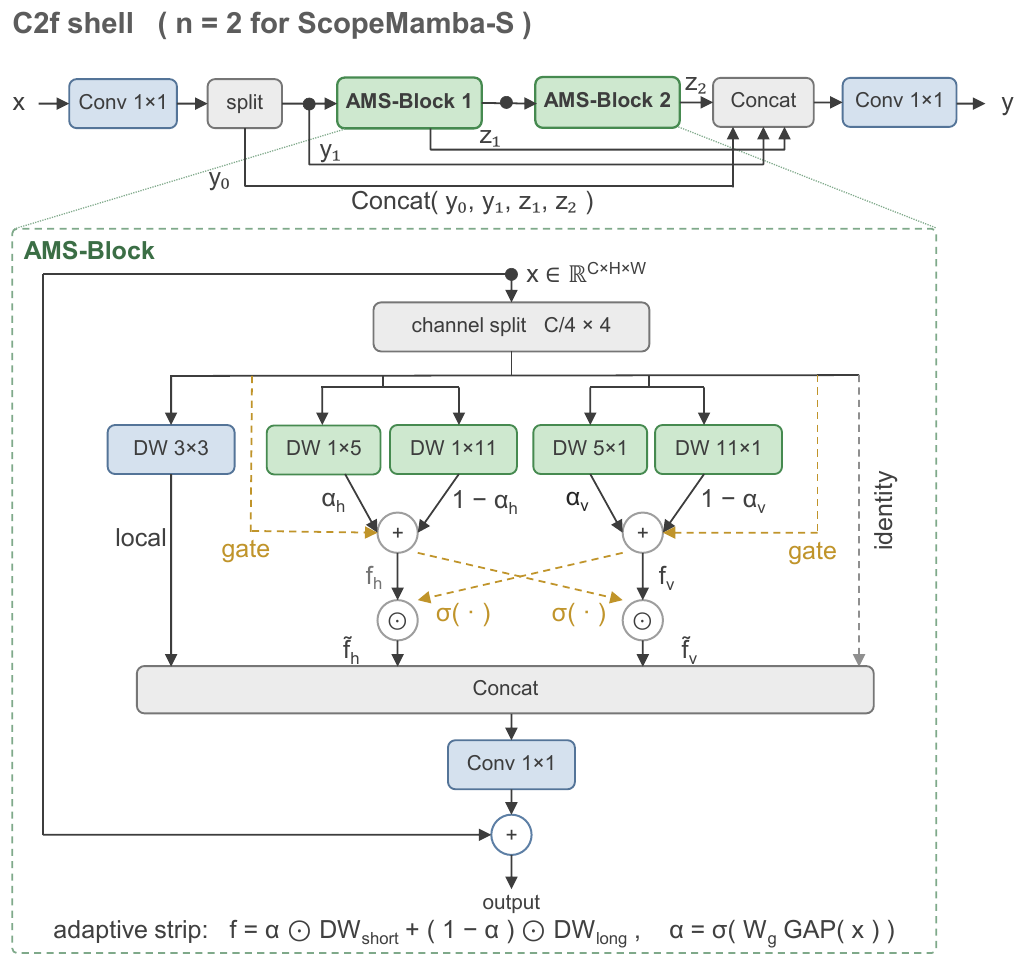}
\caption{The AMS-Block. The C2f shell is unchanged; each internal unit
splits channels four ways into a local DW $3{\times}3$ branch, adaptive
horizontal and vertical strip branches (each blending a short and a long
strip kernel under a GAP-driven gate), and an identity branch, followed
by cross-directional gating, $1{\times}1$ fusion, and a residual
connection.}
\label{fig:ams}
\end{figure}

Aerial scenes contain pronounced anisotropic structures at both the
object and contextual levels, including elongated vehicles and ships as
well as targets distributed along roads, lanes, and waterways. Such
directional patterns are not always well matched by stacks of isotropic
$3{\times}3$ kernels, which are also relatively costly on the
high-resolution P2/P3 feature maps. The AMS-Block (Adaptive Multi-scale Strip) replaces the bottleneck
inside C2f with a four-branch anisotropic unit
(Fig.~\ref{fig:ams}) and is paired with SCDown~\cite{wang2024yolov10}
downsampling, which factorizes a strided convolution into a $1{\times}1$
channel mixer followed by a stride-2 depthwise convolution.

Given an input $x\in\mathbb{R}^{C\times H\times W}$, the unit splits the
channels into four equal groups $x_1,\dots,x_4$. The first group passes
through a local depthwise $3{\times}3$ convolution,
$f_{\ell}=\mathrm{DW}_{3\times3}(x_1)$, and the fourth is an identity
shortcut. The second and third groups pass through \emph{adaptive strip
branches} in the horizontal and vertical directions. Taking the
horizontal branch as an example, a channel-wise gate blends a short and a
long strip kernel according to global content:
\begin{align}
\alpha_h&=\sigma\!\bigl(W_g\,\mathrm{GAP}(x_2)\bigr),
\label{eq:ams_gate}\\
f_h&=\alpha_h\odot \mathrm{DW}_{1\times5}(x_2)
   +(1-\alpha_h)\odot \mathrm{DW}_{1\times11}(x_2),
\label{eq:ams_strip}
\end{align}
where $\mathrm{GAP}$ is global average pooling, $W_g$ a $1{\times}1$
projection, $\sigma$ the sigmoid, and $\odot$ channel-wise
multiplication; the vertical branch $f_v$ is defined symmetrically with
$5{\times}1$ and $11{\times}1$ kernels. The two directional streams then
modulate each other through cross-directional gating,
\begin{equation}
\tilde f_h=f_h\odot\sigma\!\bigl(g_{v\to h}(f_v)\bigr),\qquad
\tilde f_v=f_v\odot\sigma\!\bigl(g_{h\to v}(f_h)\bigr),
\label{eq:ams_cross}
\end{equation}
so that, e.g., evidence of a vertical structure suppresses or sharpens
the horizontal reading at the same location. The unit output is
\begin{equation}
y=x+W_f\,\bigl[\,f_{\ell};\ \tilde f_h;\ \tilde f_v;\ x_4\,\bigr],
\label{eq:ams_fuse}
\end{equation}
with $[\cdot;\cdot]$ channel concatenation and $W_f$ a $1{\times}1$
fusion. Strip kernels touch $k$ pixels instead of $k^2$, so the unit
covers an $11\times11$ extent at close to depthwise-line cost.

The AMS-Block primarily serves as a resource-efficient feature
extraction component. When introduced into the P3/P4 stages of the HRR
baseline, it keeps mAP$_{50}$ nearly unchanged while reducing the
parameter count by approximately 24\% and GFLOPs by approximately 9\%
(Section~\ref{sec:ablation}). The resulting parameter saving provides
part of the budget required by the subsequent context pathways.

\subsection{Off-Path Zero-Gated Selective Scanning and the CGCM}
\label{sec:method_offpath}

\begin{figure*}[!t]
\centering
\includegraphics[width=\textwidth]{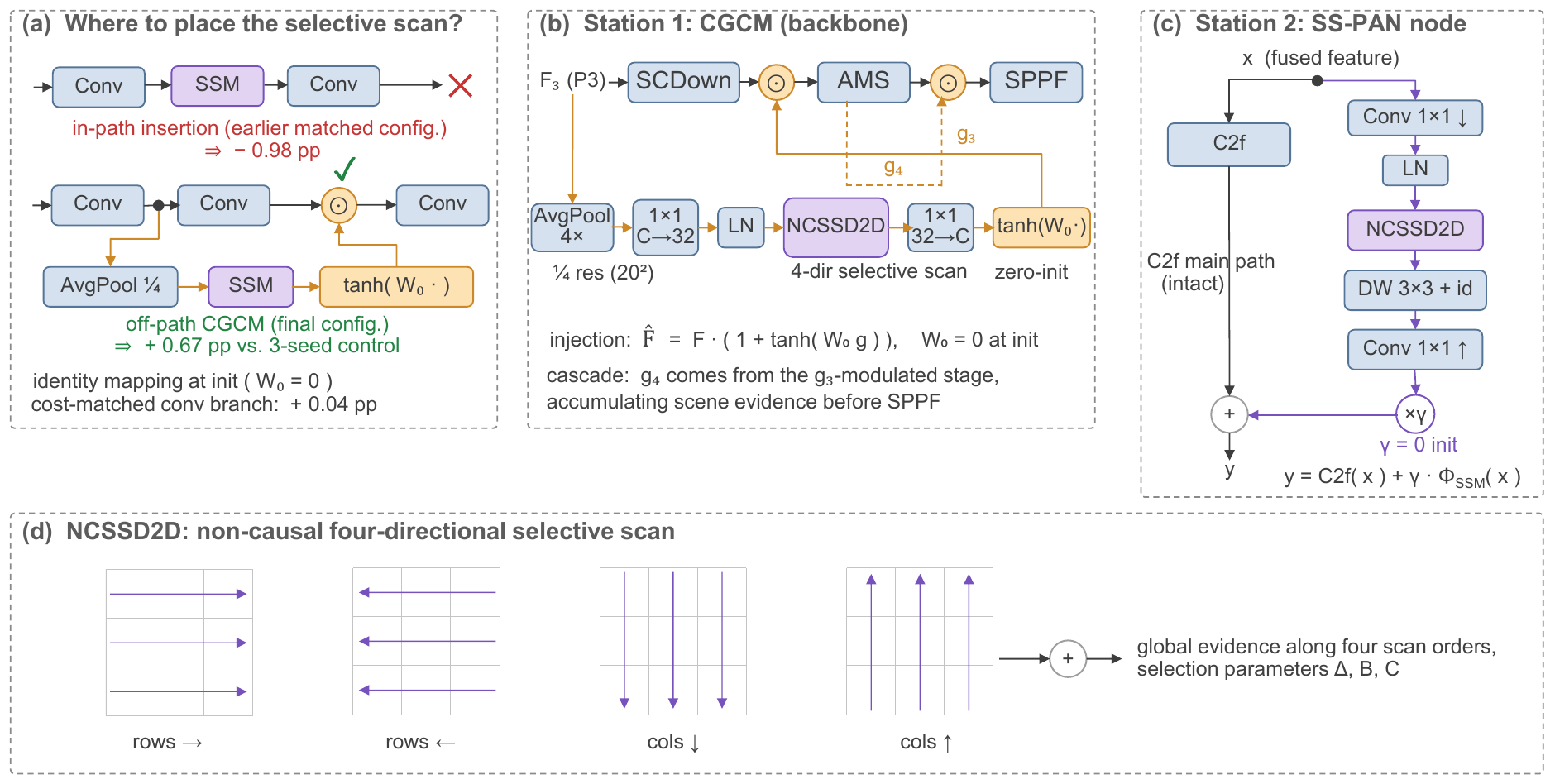}

\caption{The off-path zero-gated selective-scanning principle and its two instantiations. (a) Integration strategy matters: in an earlier matched P2/$-$P5 configuration, embedding selective scanning inside the backbone block reduces mAP$_{50}$ by 0.98 pp at unchanged GFLOPs, whereas under the final configuration the off-path CGCM improves mAP$_{50}$ by 0.67 pp over a three-seed no-branch control; a cost-matched convolutional branch at the same off-path node changes mAP$_{50}$ by only $+0.04$ pp. (b) Station~1: the Cascaded Global-Context Module in the backbone, with the $g_3$ chain shown in full and the cascaded $g_4$ computed from the already-modulated stage. (c) A bottom-up selective-scan fusion node within SS-PAN. (d) The NCSSD2D operator shared by both stations: a non-causal four-directional selective scan with linear complexity in the number of tokens.}

\label{fig:offpath}
\end{figure*}

\subsubsection{Preliminaries: selective state-space models}
A structured state-space model (SSM) maps a 1-D input $x(t)$ to an
output $y(t)$ through a latent state $h(t)\in\mathbb{R}^{N}$:
\begin{equation}
h'(t)=\mathbf{A}h(t)+\mathbf{B}x(t),\qquad
y(t)=\mathbf{C}h(t),
\label{eq:ssm_cont}
\end{equation}
which, discretized with step $\Delta$ under a zero-order hold,
\begin{equation}
\bar{\mathbf{A}}=\exp(\Delta\mathbf{A}),\quad
\bar{\mathbf{B}}=(\Delta\mathbf{A})^{-1}
\bigl(\exp(\Delta\mathbf{A})-\mathbf{I}\bigr)\Delta\mathbf{B},
\label{eq:ssm_zoh}
\end{equation}
yields the linear recurrence
\begin{equation}
h_t=\bar{\mathbf{A}}h_{t-1}+\bar{\mathbf{B}}x_t,\qquad
y_t=\mathbf{C}h_t.
\label{eq:ssm_disc}
\end{equation}
Mamba~\cite{gu2023mamba} makes $\Delta$, $\mathbf{B}$ and $\mathbf{C}$
functions of the input (\emph{selective} scanning), so each token can
decide what to retain from or contribute to the running state, while the
recurrence keeps the cost linear in sequence length. To lift the
operator to 2-D features we use a non-causal four-directional variant,
NCSSD2D: the feature map is serialized along four scan orders (rows
left$\to$right and right$\to$left, columns top$\to$bottom and
bottom$\to$top), each order is processed by a selective scan, and the
four outputs are merged (Fig.~\ref{fig:offpath}d). Every position thus
aggregates evidence from the entire map at $\mathcal{O}(HW)$ cost---the
property that makes a global-context stage affordable at all.

\subsubsection{The placement principle}
How selective scanning is integrated with the detail-carrying stream is
critical for small-object detection. In an earlier matched P2/$-$P5
configuration, inserting the scan inside the backbone block reduces
mAP$_{50}$ from 0.4949 to 0.4851 while leaving computation unchanged at
47.6 GFLOPs (Section~\ref{sec:placement}). This negative result motivates
an integration strategy in which selective scanning provides contextual
modulation without directly replacing the main convolutional stream. We
therefore impose three constraints, which together define our
\emph{off-path zero-gated selective-scanning} formulation
(Fig.~\ref{fig:offpath}a):

\begin{enumerate}
\item \textbf{Off-path, modulation only.} The scan does not replace the
main-path feature transformation. It runs on a side branch and produces
a modulation signal, allowing the convolutional stream to preserve the
fine-grained representation used for small-object localization.

\item \textbf{Reduced cost.} In the backbone, the context branch operates
on $4{\times}$-pooled, 32-channel features. The resulting
$\tfrac14$-resolution representation reduces the token count by
$16{\times}$ and limits the cumulative CGCM overhead to $+0.44$ GFLOPs
across the two backbone context stations. In the neck, where the scan
acts on the full-resolution fusion map, spatial pooling is not used;
instead, the branch operates at half the output width
(Section~\ref{sec:method_neck}).

\item \textbf{Zero-initialized gating.} The contextual signal enters
through a projection initialized at zero, making the augmented network
functionally identical to its convolutional skeleton at initialization
and allowing the context branch to be introduced progressively during
optimization.
\end{enumerate}

The gain is also not explained by the mere presence of an auxiliary
branch. At the final off-path CGCM node, a cost-matched convolutional
branch changes mAP$_{50}$ by only $+0.04$ pp and EMA by $-0.32$ pp
relative to the three-seed no-branch mean, whereas selective scanning
provides a $+0.67$ pp gain (Section~\ref{sec:placement}).

\subsubsection{Station 1: Cascaded Global-Context Module (CGCM)}
The first instantiation lives in the backbone
(Fig.~\ref{fig:offpath}b). From the P3 stage output
$F_3\in\mathbb{R}^{C\times H\times W}$ a context stream is computed as
\begin{equation}
g_3=W_{\uparrow}\,
\mathrm{NCSSD2D}\!\bigl(\mathrm{LN}\bigl(W_{\downarrow}\,
\mathrm{AvgPool}_{4}(F_3)\bigr)\bigr),
\label{eq:cgcm_ctx}
\end{equation}
where $\mathrm{AvgPool}_4$ pools by a factor of four (a $20\times20$
grid at $640^2$ input), $W_{\downarrow}$ projects to a 32-channel
working width, $\mathrm{LN}$ is LayerNorm, the scan uses state size
$d_{\mathrm{state}}{=}8$, and $W_{\uparrow}$ restores the stage width.
The context is injected multiplicatively at the entrance of the P4
stage:
\begin{equation}
\hat F = F\cdot\bigl(1+\tanh(W_0\,g)\bigr),
\label{eq:cgcm_inject}
\end{equation}
with $g$ bilinearly aligned to $F$ and the projection $W_0$
zero-initialized, so the gate spans $(0,2)$ and equals the identity at
step~0. A second context $g_4$ is then computed---by an identical
chain, from a $10\times10$ grid---\emph{from the already-modulated} P4
features and injected before SPPF. This cascade is what the module is
named for: scene evidence gathered at P3 conditions the P4 computation,
and the refreshed P4 evidence is folded in once more before the final
spatial pooling, so context accumulates along depth instead of being
applied once.

The $20\times20$ scanned grid of $g_3$ allows each contextual output to aggregate information across the pooled P3 feature map without applying selective scanning to the full-resolution backbone representation. The same formulation is applied at P4 through $g_4$, providing cascaded contextual modulation while preserving the main convolutional pathway.

\subsection{Selective-Scan PAN (SS-PAN)}
\label{sec:method_neck}

We denote the redesigned neck as the \emph{Selective-Scan PAN (SS-PAN)}. It consists of two complementary pathways: a top-down pathway that uses DySample up-sampling and directional context injection to propagate semantic information toward the high-resolution P2 feature, and a bottom-up pathway in which the P3 and P4 fusion nodes incorporate zero-gated selective-scan branches.

\subsubsection{Directional top-down pathway}
Up-sampling in the top-down pathway uses
DySample~\cite{liu2023dysample}. After the two fusion stages produce
$P_2'$, a lightweight injection enriches the P2 map with directional
semantics from $P_3'$. Specifically, $P_3'$ is up-sampled and aligned to
a 64-channel semantic map $s$, which is decomposed directionally as in
\eqref{eq:ams_cross},
\begin{equation}
s_{\mathrm{dir}}
=s_h\!\odot\!\sigma\!\bigl(g_{v\to h}(s_v)\bigr)
+s_v\!\odot\!\sigma\!\bigl(g_{h\to v}(s_h)\bigr),
\end{equation}
where $s_h=\mathrm{DW}_{1\times5}(s)$ and
$s_v=\mathrm{DW}_{5\times1}(s)$. The resulting directional semantic
signal is applied through the same zero-initialized multiplicative gate
as \eqref{eq:cgcm_inject}, yielding $P_2^{\mathrm{final}}$. This
injection follows the same general design language as the CGCM---a
zero-gated contextual correction on top of the main feature
pathway---and is treated as a lightweight neck component rather than as
a separate contribution.

\subsubsection{Bottom-Up Selective-Scan Fusion}
The bottom-up fusion nodes at P3 and P4 replace C2f with C2f-SDMamba
(Fig.~\ref{fig:offpath}c):
\begin{equation}
y=\mathrm{C2f}(x)+\gamma\,\Phi_{\mathrm{SSM}}(x),\qquad
\gamma=0 \text{ at init},
\label{eq:sspan}
\end{equation}
where the branch
$\Phi_{\mathrm{SSM}}$ compresses $x$ with a $1{\times}1$ convolution to
half the output width, normalizes, applies NCSSD2D
($d_{\mathrm{state}}{=}16$), refines with a residual depthwise
$3{\times}3$, and expands back with a $1{\times}1$ convolution. 

The node follows the same off-path formulation used in the backbone:
the C2f main path is left unchanged, while the selective-scan branch
provides a global recalibration signal for the concatenated multi-scale
features and is introduced progressively from zero. The branch is
therefore intended to complement, rather than replace, the local fusion
pathway. We place the scans at the P3 and P4 bottom-up fusion nodes;
the negative configurations in Section~\ref{sec:negative_results} show
that removing one scan node or tapering the scan width reduces accuracy,
while the tapered configuration provides no parameter advantage.

\subsection{Scale-Adaptive DFL Head}
\label{sec:method_sadfl}

\begin{figure}[!t]
\centering
\includegraphics[width=\columnwidth]{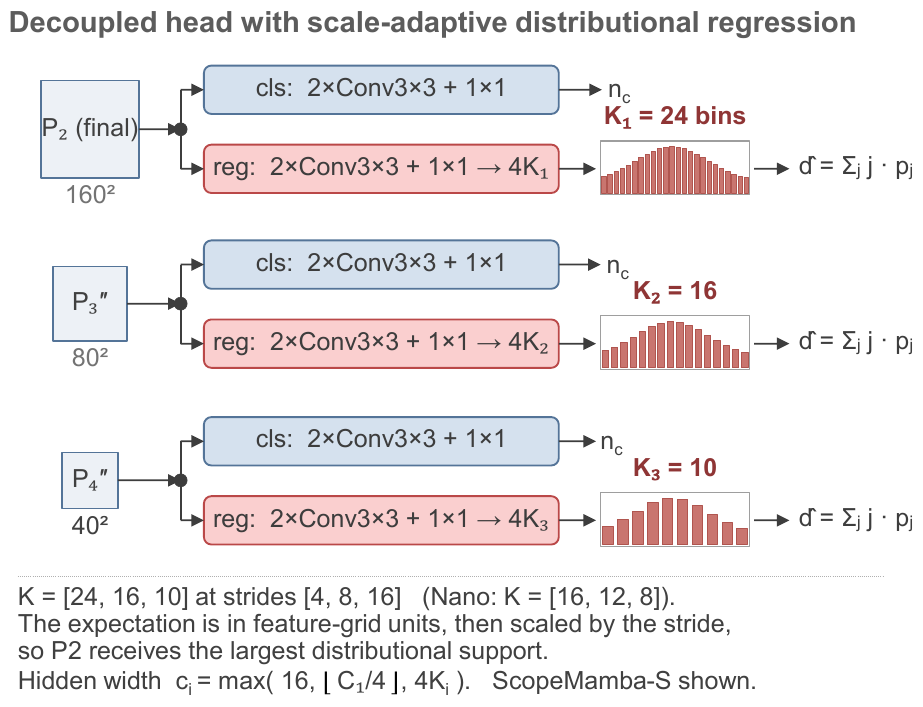}
\caption{The SA-DFL head. The decoupled head structure is standard; the
regression branch of each scale predicts a distribution over its own
number of bins ($K_1{=}24$, $K_2{=}16$, $K_3{=}10$ for strides
4/8/16), decoded by expectation. Bins are indexed in stride units, so
the per-scale bin budget determines the supported regression range
rather than changing the spacing between adjacent bins.}
\label{fig:sadfl}
\end{figure}

Distribution Focal Loss~\cite{li2020generalized} regresses each box side
as the expectation of a discrete distribution over $K$ bins indexed in
stride units. Standard YOLOv8 uses a common $K{=}16$ at all detection
levels. After the P2/P3/P4 reallocation, however, the same image-space
offset corresponds to different numbers of feature-grid units at
different strides. We therefore assign a scale-specific bin budget
$K_i$ to each regression branch. Changing $K_i$ adjusts the
distributional support available at that level; under the decoding used
here, it does not reduce the spacing between adjacent bin indices. The
regression branch of scale $i$ predicts logits
$z\in\mathbb{R}^{4K_i}$ and decodes each side distance as
\begin{equation}
\hat d=\sum_{j=0}^{K_i-1} j\cdot p_j,\qquad
p=\mathrm{softmax}(z_{\mathrm{side}}),
\label{eq:sadfl_decode}
\end{equation}
with per-scale output channels $n_o^{(i)}=n_c+4K_i$ and regression
hidden width
\begin{equation}
c_i=\max\bigl(16,\ \lfloor C_1/4\rfloor,\ 4K_i\bigr),
\label{eq:sadfl_width}
\end{equation}
where $C_1$ is the first head input width.

Because the regression-branch width scales with $4K_i$, changing the
per-scale bin allocation simultaneously reallocates regression capacity
across feature levels. We therefore attribute the observed improvement
to the joint scale-adaptive regression design---distributional support
together with regression capacity---rather than to finer bin spacing
alone. Training uses a scale-indexed DFL term, with each detection level
supervised using its corresponding $K_i$, alongside the standard CIoU
and BCE classification losses; no additional inputs or quality branches
are introduced. The primary configuration uses
$K{=}[24,16,10]$ for strides $[4,8,16]$
(Fig.~\ref{fig:sadfl}). This modification adds only 0.008M parameters
while providing a positive accuracy gain in the cumulative ablation
(Section~\ref{sec:bins}).

\subsection{Model Scaling}
\label{sec:method_scaling}

\begin{table}[!t]
\centering
\caption{ScopeMamba-YOLO variants. All four share the same topology; only the compound scaling factors (depth $d$, width
$w$) and, for the Nano tier, the SA-DFL bin list differ. Params and
GFLOPs are reported at $640\times640$ input under the 10-class
VisDrone configuration.}
\label{tab:variants}
\setlength{\tabcolsep}{2.5pt}
\footnotesize
\begin{tabular}{lccccc}
\toprule
Variant & $d$ & $w$ & $K$ (P2/P3/P4) & Params & GFLOPs \\
\midrule
ScopeMamba-N & 0.33 & 0.25  & [16, 12, 8]  & 1.01M & 15.87 \\
ScopeMamba-S & 0.33 & 0.50  & [24, 16, 10] & 3.57M & 53.69 \\
ScopeMamba-M & 0.67 & 0.625 & [24, 16, 10] & 6.48M & 92.17 \\
ScopeMamba-L & 0.67 & 0.75  & [24, 16, 10] & 9.17M & 127.78 \\
\bottomrule
\end{tabular}
\end{table}

ScopeMamba-YOLO follows YOLOv8 compound scaling: nominal per-layer
channel widths are multiplied by $w$ and rounded to multiples of eight,
and the repeat counts of C2f-family blocks are scaled by $d$. All four
variants in Table~\ref{tab:variants} instantiate the same topology; no
module is added, removed, or rewired across tiers. The single
hyperparameter that does change is the Nano bin list. By
\eqref{eq:sadfl_width}, the regression hidden width is dominated by
$4K_i$ once the model is narrow: at Nano width ($C_1{=}64$), retaining
$K{=}[24,16,10]$ would pin the regression branches at widths
$[96,64,40]$---wider than their own inputs and immune to width scaling.
Shrinking to $K{=}[16,12,8]$ restores proportional scaling
($[64,48,32]$), mirroring how YOLOv8 itself caps head widths per tier;
Section~\ref{sec:bins} shows that this choice performs best among the
tested Nano-scale bin allocations. We report the S variant as the primary model throughout
the experiments.

\section{EXPERIMENTS}

In this section, we present comprehensive experimental results to validate the effectiveness of the proposed method. Benchmark evaluations are conducted on two representative object detection datasets: VisDrone-2019 \cite{zhu2022detection} and AI-TOD \cite{wang2021tiny}. We adopt YOLOv8s as our baseline detector.

To quantify detection performance, we adopt the standard mean average precision (mAP) metric, including $mAP_{50}$ and $mAP_{50-95}$, as well as the number of parameters and FLOPs to evaluate the model complexity.

All experiments are conducted on Ubuntu using PyTorch and an NVIDIA
RTX 4090 GPU. Input images are uniformly resized to
$640\times640$. Models are optimized using stochastic gradient descent
(SGD) with an initial learning rate of 0.01, momentum of 0.937, and
weight decay of 0.0005, with a batch size of 8. The maximum number of
training epochs is 500 for VisDrone-2019 and 400 for AI-TOD, and early
stopping with a patience of 30 epochs is applied uniformly. No
pre-trained parameters are used in any experiment.

\subsection{Experimental Dataset}

\textbf{VisDrone-2019:} This dataset \cite{zhu2022detection} contains 6,471 training images, 548 validation images captured by drones, and 1,610 test images. Our ScopeMamba results, controlled baselines, and ablation experiments are evaluated on the official 548-image validation split. The resolution of these images is approximately 1000--1500 pixels. These images are annotated with bounding boxes across ten categories: pedestrian, bicycle, tricycle, people, truck, car, bus, van, motor, and awning-tricycle. Object detection on this dataset remains challenging due to heavy occlusion, extreme scale variation, uneven spatial distribution, and the distinct lack of large targets. To provide a statistical foundation for our architectural design, Fig.~\ref{fig:bbox_area_hist} illustrates the bounding box size distribution of the dataset. The overwhelming concentration of objects in the extremely small regimes explicitly motivates our high-resolution reallocation strategy.

\textbf{AI-TOD:} This benchmark \cite{wang2021tiny} is curated for tiny
object detection in aerial images, comprising 28,036 images and 700,621
instances across eight categories. We adopt the official partition of
11,214 training and 2,804 validation images and conduct evaluation on
the official validation split. Images are uniformly resized from their
native $800\times800$ resolution to $640\times640$. AI-TOD is
characterized by extremely small and densely distributed objects, with
a mean target size of approximately 12.8 pixels. In addition to
$mAP_{50}$, $mAP_{75}$, and $mAP_{50-95}$, we report scale-specific
metrics for very tiny (2--8 pixels, $AP_{vt}$), tiny (8--16 pixels,
$AP_t$), and small (16--32 pixels, $AP_s$) objects. To accommodate the
high target density, the maximum number of detections per image is set
to 1,500.

\subsection{Comparisons with State-of-the-Art Methods}
We compare ScopeMamba-YOLO with the baseline and recent aerial-object detectors. For VisDrone-2019, the results of comparison methods are taken from the corresponding cited publications. For AI-TOD, the comparison methods are reproduced from their publicly released implementations under the unified training and evaluation protocol described above.

\subsubsection{VisDrone-2019}

\begin{figure}[t]
\centering
\includegraphics[width=\linewidth]{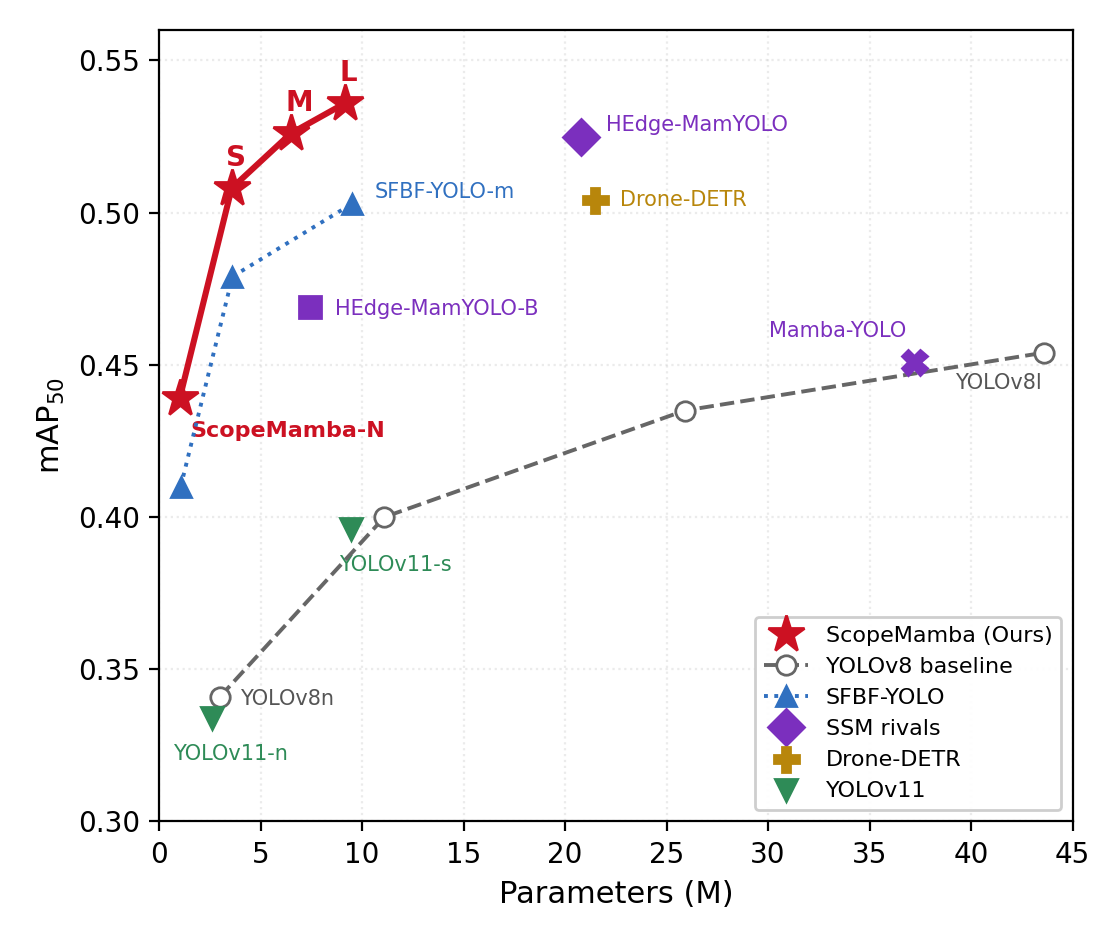}
\caption{Accuracy ($mAP_{50}$) versus parameter count on the
VisDrone-2019 dataset. Among the compared methods, the ScopeMamba
variants occupy a favorable region of the accuracy--parameter plane
across different model scales.}
\label{fig:acc_png}
\end{figure}

Table~\ref{tab:visdrone_comparison} compares ScopeMamba-YOLO with the
baseline and recent aerial-object detectors on VisDrone-2019.
Fig.~\ref{fig:acc_png} additionally visualizes mAP$_{50}$ against
parameter count. Across the evaluated scales, the ScopeMamba variants
show a favorable accuracy--parameter trade-off relative to the compared
methods.

\begin{table*}[t]
\centering
\caption{Comprehensive Performance Comparison on the VisDrone-2019 Dataset. 
Models are categorized by scale (Nano, Small, Medium, Large) from top to bottom. 
Best results are highlighted in \textbf{bold}, and second-best results are \underline{underlined}.}
\label{tab:visdrone_comparison}
\small
\begin{tabular*}{\textwidth}{@{\extracolsep{\fill}}lcccc}
\toprule
Model & Params (M) & GFLOPs & mAP$_{50}$ & mAP$_{50-95}$ \\

\midrule
YOLOv8n~\cite{jocher2023yolov8} & 3.00 & \underline{8.1} & 0.341 & 0.195 \\
YOLOv10-n~\cite{wang2024yolov10} & 2.71 & 8.4 & 0.333 & 0.190 \\
YOLOv11-n~\cite{khanam2024yolov11} & 2.59 & \textbf{6.5} & 0.334 & 0.191 \\
SFBF-YOLO-n~\cite{hua2026sfbf} & \underline{1.06} & 13.7 & \underline{0.410} & \underline{0.243} \\
\textbf{ScopeMamba-N (Ours)} & \textbf{1.01} & 15.87 & \textbf{0.439} & \textbf{0.266} \\

\midrule
YOLOv8s & 11.10 & 28.7 & 0.400 & 0.238 \\
YOLOv10-s & 8.10 & 24.8 & 0.391 & 0.229 \\
YOLOv11-s & 9.43 & \underline{21.6} & 0.396 & 0.232 \\
HEdge-MamYOLO-B~\cite{wang2026hedge} & 7.42 & 42.8 & 0.469 & 0.289 \\
SFBF-YOLO-s & \underline{3.61} & 42.6 & \underline{0.479} & \underline{0.292} \\
LMF-UAV(s)~\cite{yang2025lmf} & 6.30 & \textbf{17.6} & 0.364 & 0.213 \\
PC-YOLO11s~\cite{wang2025pcyolo11s} & 7.10 & - & 0.438 & 0.263 \\
UAV-YOLO~\cite{wang2023uavyolov8} & 10.30 & - & 0.470 & \underline{0.292} \\
DMFF-YOLO~\cite{qiu2024dmffyolo} & 7.65 & - & 0.447 & 0.272 \\
BDH-YOLO~\cite{sui2024bdhyolo} & 9.39 & - & 0.429 & 0.262 \\
\textbf{ScopeMamba-S (Ours)} & \textbf{3.57} & 53.69 & \textbf{0.508} & \textbf{0.314} \\

\midrule
YOLOv8m & 25.90 & 79.1 & 0.435 & 0.263 \\
HEdge-MamYOLO & 20.80 & 151.2 & \underline{0.525} & \underline{0.328} \\
SFBF-YOLO-m & 9.52 & 100.6 & 0.503 & 0.307 \\
SOD-YOLO-m~\cite{li2024sodyolo} & \textbf{6.30} & \textbf{65.8} & 0.485 & 0.295 \\
PVswin-YOLOv8-s~\cite{tahir2024pvswin} & 21.60 & - & 0.433 & 0.264 \\
Drone-DETR~\cite{wang2025dronedetr} & 21.47 &  \underline{73.9} & 0.504 & 0.311 \\

\textbf{ScopeMamba-M (Ours)} & \underline{6.48} & 92.17 & \textbf{0.526} & \textbf{0.332} \\

\midrule
YOLOv8l & 43.60 & 164.9 & 0.454 & 0.279 \\
YOLOv9-c~\cite{wang2024yolov9} & 50.90 & 237.8 & 0.481 & 0.299 \\
Mamba-YOLO~\cite{wang2025mamba} & 37.17 & 94.3 & 0.451 & 0.276 \\
SOD-YOLO-l & 17.60 & 167.0 & 0.515 & \underline{0.320} \\
LMF-UAV(l) & \underline{14.70} & \textbf{61.8} & 0.411 & 0.246 \\
EFA-Net~\cite{liu2025efanet} & 37.40 & 108.2 & \underline{0.516} & 0.296 \\
YOLO-DCTI~\cite{min2023yolodcti} & 37.60 & - & 0.498 & 0.274 \\
RemDet-L~\cite{li2025remdet} & - & \underline{67.4} & 0.473 & - \\
RemDet-X & - & 114.0 & 0.483 & - \\
CEASC(GFL V1)~\cite{du2023adaptive} & - & 150.0 & 0.507 & - \\
CEASC(FSAF) & - & 153.0 & 0.489 & - \\
CEASC(Faster-RCNN) & - & 133.0 & 0.434 & - \\
\textbf{ScopeMamba-L (Ours)} & \textbf{9.17} & 127.78 & \textbf{0.536} & \textbf{0.337} \\

\bottomrule
\end{tabular*}
\end{table*}

\textbf{Comparison with Baseline YOLO Models:}
ScopeMamba consistently improves detection accuracy over the YOLOv8
entries across the evaluated model scales while using substantially
fewer parameters. ScopeMamba-N improves mAP$_{50}$ by 9.8 pp
(0.439 versus 0.341) with 1.01M parameters, compared with 3.00M for
YOLOv8n. At the Small scale, ScopeMamba-S improves mAP$_{50}$ by
10.8 pp over YOLOv8s while reducing the parameter count from 11.10M
to 3.57M. ScopeMamba-L reaches 0.536 mAP$_{50}$ with 9.17M parameters,
compared with 0.454 and 43.60M parameters for YOLOv8l.

\textbf{Comparison with SSM-based Detectors:}
Compared with recent SSM-based detectors, ScopeMamba achieves a
favorable accuracy--parameter trade-off. Mamba-YOLO reports
0.451 mAP$_{50}$ with 37.17M parameters, whereas ScopeMamba-S reaches
0.508 with 3.57M parameters. HEdge-MamYOLO reaches 0.525 mAP$_{50}$
with 20.80M parameters and 151.2 GFLOPs; ScopeMamba-M reaches 0.526
with 6.48M parameters and 92.17 GFLOPs.

\textbf{Comparison with Other UAV-Specific SOTAs:}
ScopeMamba also compares favorably with specialized detectors designed
for aerial imagery. ScopeMamba-M reaches 0.526 mAP$_{50}$ and 0.332
mAP$_{50-95}$ with 6.48M parameters, exceeding the reported accuracy
of Drone-DETR, SFBF-YOLO-m, and SOD-YOLO-l while using fewer parameters
than most of these alternatives. These comparisons further demonstrate
the favorable accuracy--parameter trade-off of the proposed model.

Fig.~\ref{fig:acc_png} visualizes the accuracy--parameter trade-off
among the compared detectors. Relative to the corresponding YOLOv8
entries, the ScopeMamba variants consistently achieve higher
mAP$_{50}$ with fewer parameters. Compared with recent SSM-based
detectors, ScopeMamba-S reaches 0.508 mAP$_{50}$ with 3.57M parameters,
whereas Mamba-YOLO reports 0.451 with 37.17M parameters; ScopeMamba-M
reaches 0.526 with 6.48M parameters, compared with 0.525 and 20.80M
parameters for HEdge-MamYOLO. These results indicate a favorable
accuracy--parameter trade-off for the ScopeMamba series among the
methods included in this comparison. Evidence for the proposed
integration strategy itself is provided separately by the controlled
experiments in Section~\ref{sec:placement}.

\subsubsection{AI-TOD}

Table~\ref{tab:comparison_results} compares ScopeMamba-YOLO with representative detectors on AI-TOD in terms of overall accuracy, size-specific AP/AR, model complexity, and computational cost.

\begin{table*}[t]

\centering

\caption{Comparison of Different Methods on the AI-TOD Dataset.}

\label{tab:comparison_results}
\small
\setlength{\tabcolsep}{4pt}
\begin{tabular}{lllllllllllll}

\toprule

\textbf{Model}
& \textbf{Params(M)}
& \textbf{Resolution}
& \textbf{mAP$_{50-95}$}
& \textbf{mAP$_{50}$}
& \textbf{mAP$_{75}$}
& $AP^{vt}$
& $AP^{t}$
& $AP^{s}$
& $AR^{vt}$
& $AR^{t}$
& $AR^{s}$
& \textbf{GFLOPs} \\

\midrule

\begin{tabular}[c]{@{}l@{}}
ALSS-YOLO-\\ S~\cite{he2024alssyolo}
\end{tabular}
& \underline{2.18}
& $640 \times 640$
& 12.90
& 31.77
& 7.55
& 2.22
& 12.83
& 19.94
& 3.67
& 22.98
& 30.45
& \textbf{8.50} \\

\begin{tabular}[c]{@{}l@{}}
ALSS-YOLO-\\ M
\end{tabular}
& 2.74
& $640 \times 640$
& 12.94
& 32.83
& 7.83
& 2.40
& 12.42
& 20.20
& 4.92
& 22.42
& 31.33
& \underline{10.40} \\

\textbf{ScopeMamba-N}
& \textbf{1.01}
& $640 \times 640$
& \textbf{19.71}
& \textbf{45.24}
& \textbf{13.70}
& \textbf{6.77}
& \textbf{20.64}
& \textbf{24.33}
& \textbf{12.78}
& \textbf{37.88}
& \textbf{38.41}
& 15.86 \\

\midrule

L-FFCA-YOLO~\cite{zhang2024ffca}
& \underline{5.06}
& $640 \times 640$
& \underline{22.83}
& \underline{51.80}
& 16.34
& \textbf{10.70}
& \underline{24.60}
& \underline{29.85}
& \textbf{25.53}
& \textbf{45.22}
& 43.56
& 37.20 \\

YOLOv9-S~\cite{wang2024yolov9}
& 7.29
& $640 \times 640$
& 20.33
& 47.54
& 14.16
& 7.52
& 20.13
& 27.96
& 13.65
& 33.68
& 39.17
& 27.08 \\

YOLOv10s~\cite{wang2024yolov10}
& 8.07
& $640 \times 640$
& 19.78
& 45.54
& 13.95
& 6.64
& 20.38
& 26.97
& 15.09
& 35.89
& 40.43
& \underline{24.80} \\

MAF-YOLO-S~\cite{yang2024mafyolo}
& 8.51
& $640 \times 640$
& 22.89
& 49.81
& 18.24
& 5.12
& 23.74
& 32.09
& 14.28
& 39.06
& 45.15
& 25.21 \\

YOLO11s~\cite{khanam2024yolov11}
& 9.43
& $640 \times 640$
& 20.25
& 46.87
& 14.66
& 9.03
& 20.51
& 27.78
& 13.34
& 33.49
& 40.45
& \textbf{21.60} \\

YOLOv8-S~\cite{jocher2023yolov8}
& 11.14
& $640 \times 640$
& 20.80
& 48.45
& 14.69
& 5.73
& 20.97
& 29.10
& 9.56
& 34.41
& 40.65
& 28.45 \\

\textbf{ScopeMamba-S}
& \textbf{3.56}
& $640 \times 640$
& \textbf{23.67}
& \textbf{52.35}
& \underline{17.69}
& \underline{10.30}
& \textbf{25.55}
& 29.51
& \underline{21.89}
& \underline{43.66}
& \textbf{45.80}
& 53.67 \\

\midrule

YOLOv8m
& 25.86
& $640 \times 640$
& 23.51
& 52.68
& \underline{18.53}
& \textbf{13.30}
& 23.46
& 31.24
& 20.59
& 38.05
& 42.38
& \textbf{78.70} \\

\begin{tabular}[c]{@{}l@{}}
TPH-\\ YOLOv5-L~\cite{zhu2021tph}
\end{tabular}
& 41.55
& $640 \times 640$
& \underline{25.17}
& \underline{56.65}
& 18.16
& \underline{12.88}
& 27.14
& \underline{33.15}
& 21.72
& 42.06
& 42.61
& 108.00 \\

\textbf{ScopeMamba-M}
& \textbf{6.48}
& $640 \times 640$
& 24.87
& 54.68
& 18.22
& 9.43
& \textbf{28.30}
& 31.98
& \underline{22.08}
& \underline{43.33}
& \underline{46.05}
& \underline{92.15} \\

\textbf{ScopeMamba-L}
& \underline{9.17}
& $640 \times 640$
& \textbf{26.64}
& \textbf{56.86}
& \textbf{20.91}
& 11.03
& \underline{27.75}
& \textbf{34.29}
& \textbf{23.00}
& \textbf{43.76}
& \textbf{47.87}
& 127.76 \\

\bottomrule

\end{tabular}

\end{table*}

\begin{figure}[htbp]
\centering
\includegraphics[width=\linewidth]{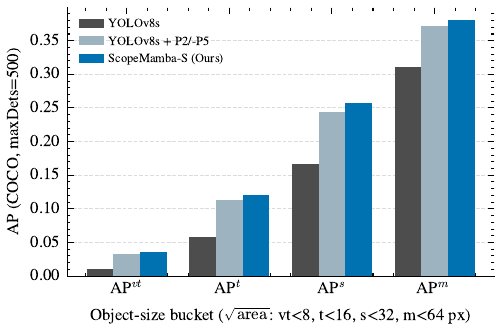}
\caption{Scale-specific performance comparison on the AI-TOD dataset.
ScopeMamba-S shows its largest improvements over YOLOv8-S in the
very-tiny and tiny object regimes. The figure reports model-level
differences rather than a per-module ablation.}
\label{fig:ap_buckets}
\end{figure}

Fig.~\ref{fig:ap_buckets} further breaks down the AI-TOD results by
object size. At the S scale, ScopeMamba-S improves over YOLOv8-S by
4.57 pp on $AP^{vt}$ (10.30 versus 5.73) and 4.58 pp on $AP^{t}$
(25.55 versus 20.97), whereas the improvement on $AP^{s}$ is more
modest at 0.41 pp (29.51 versus 29.10). The larger gains in the
very-tiny and tiny regimes are consistent with the objective of
reallocating spatial resolution toward small targets while preserving
broader contextual information. We do not attribute these bucket-level
gains to individual modules, since the module-wise ablations in this
study are conducted on VisDrone-2019 rather than AI-TOD.

\begin{figure*}[htbp]
\centering
\includegraphics[width=\textwidth]{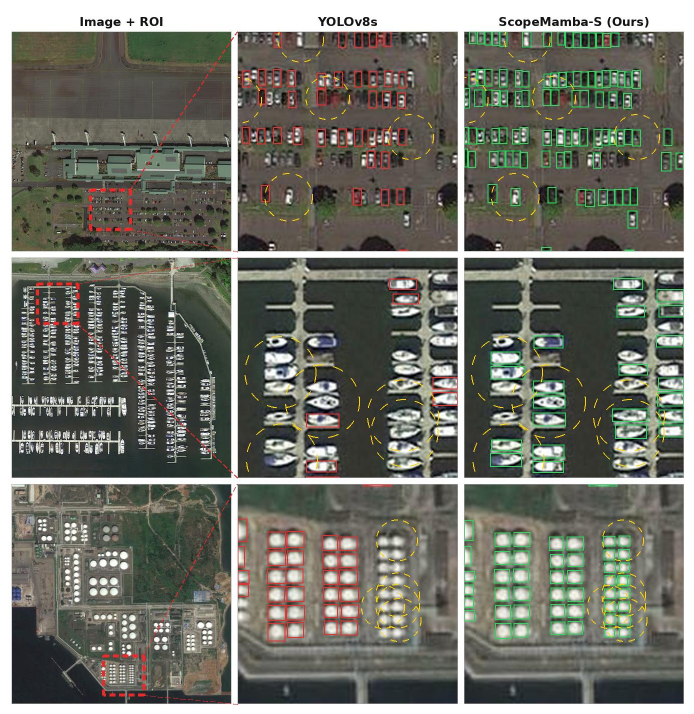}
\caption{Qualitative detection results on the AI-TOD validation set.
Yellow dashed circles indicate representative ground-truth targets
missed by YOLOv8s but detected by ScopeMamba-S in dense or very-tiny
object regions.}
\label{fig:aitod_qualitative}
\end{figure*}

Fig.~\ref{fig:aitod_qualitative} provides representative qualitative
comparisons on AI-TOD scenes containing dense and very small targets.

\subsection{Integration Strategy for Selective Scanning}
\label{sec:placement}

\textbf{The Contextual Reach Dilemma.}
A common small-object detection design adds a high-resolution stride-4
(P2) detection level while removing the stride-32 (P5) stage to control
model size. This allocation improves the representation of tiny targets
but also removes the deepest feature stage. Our ERF analysis below shows
that the resulting P2/$-$P5 configuration has substantially lower
peripheral spatial support than the original YOLOv8s. This motivates
the question examined in this section: how can broader contextual
support be reintroduced without restoring the full P5 pathway?

\textbf{Main-Path vs. Off-Path Integration.}
Recent Mamba-based detectors commonly place SSM blocks directly on the
main backbone or neck pathway. We examine this choice using two
complementary controls from different stages of the ablation lineage.
First, in an earlier P2/$-$P5 configuration with the default head,
the control configuration without the selective-scan branch achieves
0.4949 mAP$_{50}$ with 3.44M parameters and 47.6 GFLOPs. Replacing the corresponding identity branch in the matched backbone
block with selective scanning yields 0.4851 mAP$_{50}$ with 3.45M
parameters at the same 47.6 GFLOPs. This matched in-path control therefore yields a $-0.98$ pp change. Second, under the
final configuration, removing the off-path CGCM produces a three-seed
mean of 0.5011, whereas the complete model reaches 0.5078, corresponding
to a $+0.67$ pp gain. Because the in-path and off-path controls were
measured at different stages of the model lineage, we do not subtract
these two effects or interpret them as a 1.65-pp placement-only gain.
Instead, they provide complementary evidence that direct in-path
insertion is unfavorable in the matched early-stage control, whereas
the proposed off-path formulation contributes positively under the final
configuration.

\begin{figure}[htbp]
\centering
\includegraphics[width=\linewidth]{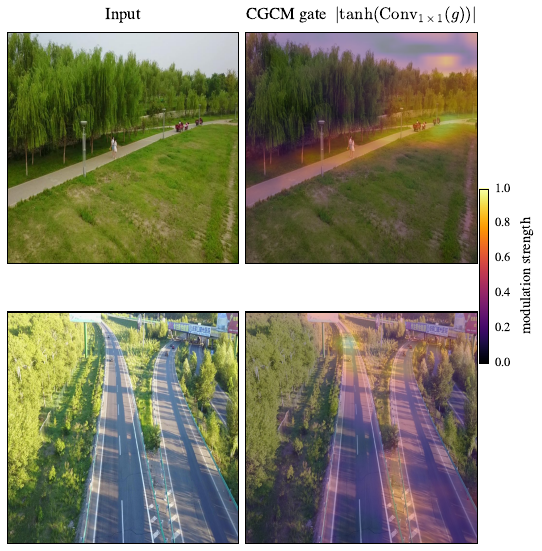}
\caption{Visualization of the CGCM gate modulation strength
$|\tanh(\mathrm{Conv}_{1\times1}(g))|$. Broad spatial responses extend
beyond object-local regions, consistent with the scene-level contextual
role of the off-path CGCM branch.}
\label{fig:cgcm_modulation}
\end{figure}

To inspect the spatial behavior of the off-path modulation,
Fig.~\ref{fig:cgcm_modulation} visualizes the magnitude of the CGCM
gate. The response is distributed over broad scene structures rather
than being confined to localized object regions, which is consistent
with the intended role of CGCM as a scene-level context modulator.

At the off-path position, the operator identity also matters. Relative
to the three-seed no-branch mean of 0.5011, a cost-matched convolutional
branch changes mAP$_{50}$ by only $+0.04$ pp and EMA by $-0.32$ pp,
whereas selective scanning improves it by $+0.67$ pp. These results are
interpreted against the observed run-to-run variation of the control
rather than as a formal significance test.

\begin{table*}[htbp]
\centering
\caption{Placement and operator controls on VisDrone-2019 at the Small
(S) scale. The in-path pair was evaluated at an earlier P2/$-$P5 stage
with the default head, whereas the off-path controls were evaluated on
the final configuration. Therefore, deltas are interpreted within each
group rather than across groups.}
\label{tab:operator_bracket}
\footnotesize
\setlength{\tabcolsep}{4pt}

\begin{tabular}{llcccc}
\toprule
\textbf{Configuration}
& \textbf{Branch}
& \textbf{Params (M)}
& \textbf{GFLOPs}
& \textbf{mAP$_{50}$}
& \textbf{$\Delta$ vs. Control} \\
\midrule

\multicolumn{6}{l}{\textit{In-path control: earlier P2/$-$P5 configuration with default head}} \\

Backbone block
& No scan branch
& 3.44
& 47.6
& 0.4949
& -- \\

Backbone block
& In-path selective scan
& 3.45
& 47.6
& 0.4851
& $-0.98$ pp \\

\midrule

\multicolumn{6}{l}{\textit{Off-path CGCM control: final configuration}} \\

Backbone CGCM
& None (3-seed mean)
& 3.39
& 53.25
& 0.5011
& -- \\

Backbone CGCM
& Convolution
& 3.57
& 53.69
& 0.5015
& $+0.04$ pp \\

Backbone CGCM
& EMA~\cite{ouyang2023efficient}
& 3.55
& 53.68
& 0.4979
& $-0.32$ pp \\

Backbone CGCM
& \textbf{Selective scan (ours)}
& 3.57
& 53.69
& \textbf{0.5078}
& $\mathbf{+0.67}$ pp \\

\midrule

\multicolumn{6}{l}{\textit{Off-path SS-PAN scan-branch control: final configuration}} \\

SS-PAN scan branches
& None
& 3.35
& 52.61
& 0.4969
& -- \\

SS-PAN scan branches
& EMA~\cite{ouyang2023efficient}
& 3.46
& 53.28
& 0.5013
& $+0.44$ pp \\

SS-PAN scan branches
& \textbf{Selective scan (ours)}
& 3.57
& 53.69
& \textbf{0.5078}
& $\mathbf{+1.09}$ pp \\

\bottomrule

\multicolumn{6}{l}{\footnotesize
\textit{Note:} The three no-CGCM runs are 0.5048, 0.5004, and 0.4980
(mean 0.5011; standard deviation 0.34 pp).}

\end{tabular}
\end{table*}

\textbf{The Operator Bracket: Why Selective Scan?}
We next test whether the positive off-path result can be explained
simply by adding an auxiliary branch. At the CGCM node, we replace the
selective scan with a cost-matched convolutional branch and with
Efficient Multi-scale Attention (EMA)~\cite{ouyang2023efficient}, while
retaining the same off-path wrapping and gating strategy.

As shown in Table~\ref{tab:operator_bracket}, the three-seed no-branch
control has a mean mAP$_{50}$ of 0.5011 with a standard deviation of
0.34 pp. The convolutional branch reaches 0.5015 ($+0.04$ pp) and EMA
reaches 0.4979 ($-0.32$ pp), both within the observed seed variation of
the control, whereas selective scanning reaches 0.5078 ($+0.67$ pp).
At the selective-scan nodes within SS-PAN, selective scanning likewise
outperforms both the no-branch configuration and the EMA substitution. Among the operators
tested at these nodes, the selective scan therefore provides the
strongest and most consistent gain, supporting the use of a
long-range state-space operator rather than attributing the improvement
to auxiliary branch capacity alone.

\textbf{Broadening Reach.}
The two context mechanisms operate under different computational
regimes: CGCM performs selective scanning on spatially pooled backbone
features, whereas the selective-scan branches within SS-PAN operate at
native neck resolution with reduced channel width. We next examine
whether their combined use broadens the spatial support weakened by
removing the P5 stage.

To visually and quantitatively corroborate this, we probe the effective receptive field (ERF) of the shared high-resolution P3 (Stride-8) detection cell. For each of the 200 VisDrone validation images uniformly resized to $640 \times 640$, we back-propagate the channel-summed response of the center cell of the P3 feature map to the input layer. We accumulate the absolute input-gradient magnitude across channels, and the final ERF map, denoted as $S(x,y)$, is averaged over the 200 images. To quantify peripheral spatial support in the effective receptive field, we define the \textit{Peripheral Energy Ratio} (PER) as the fraction of total ERF mass lying strictly outside the central $50\% \times 50\%$ ($320 \times 320$) region: \begin{equation}
\mathrm{PER}
=
1-
\frac{\sum_{(x,y)\in C} S(x,y)}
{\sum_{(x,y)} S(x,y)},
\label{eq:per}
\end{equation}
where $C$ denotes the centered half-size bounding box. A higher PER indicates broader gradient-based spatial sensitivity to
the image periphery, whereas a lower value indicates a more
center-concentrated effective receptive field. 

\begin{figure*}[t]
\centering
\includegraphics[width=\textwidth]{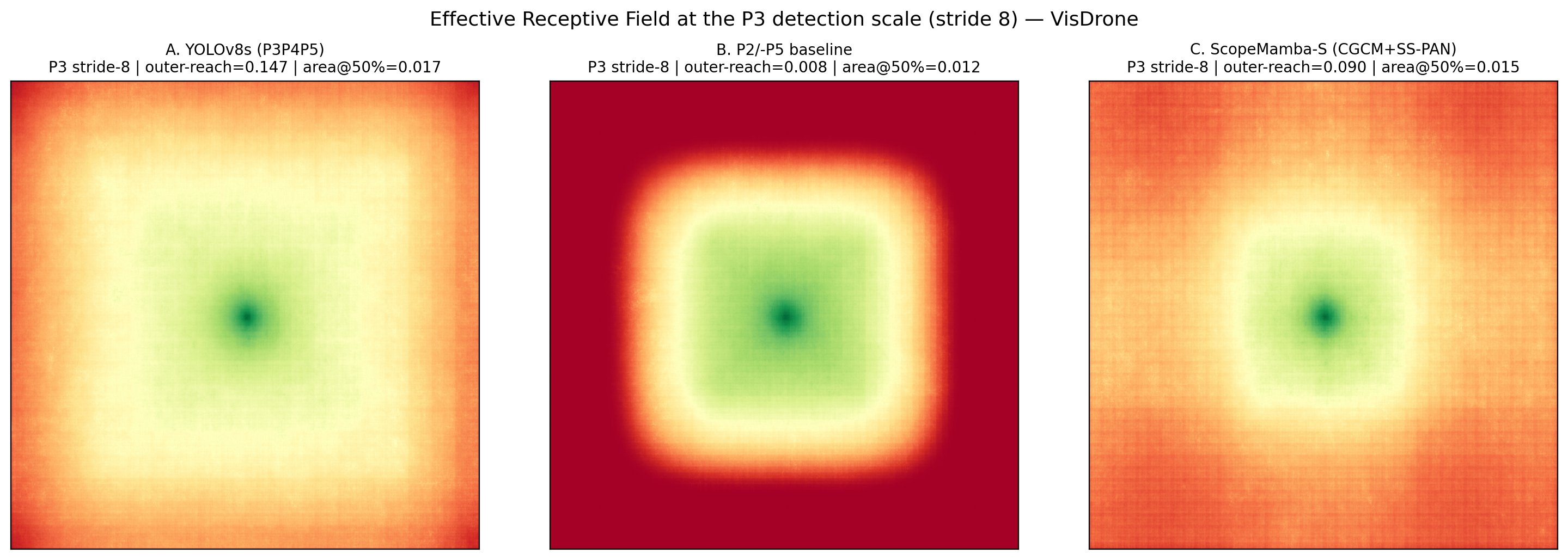} 

\caption{Effective Receptive Field (ERF) measured at the shared P3
detection feature (Stride 8). (A) The original YOLOv8s exhibits a PER
of 0.147. (B) Removing P5 while introducing the P2 detection level
concentrates the ERF near the center and reduces PER to 0.008.
(C) ScopeMamba-S increases PER to 0.090 through the off-path context
pathway, an $11\times$ expansion over the P2/$-$P5 baseline and
approximately 61\% of the peripheral reach measured in the unpruned
YOLOv8s.}

\label{fig:erf_p3}
\end{figure*}

As depicted in Fig.~\ref{fig:erf_p3}B, removing the deep P5 stage
substantially concentrates the ERF around the image center and weakens
its peripheral support. The corresponding PER decreases to 0.008,
indicating that only a small fraction of the total ERF energy remains
outside the central half-size region.

Conversely, with the off-path integration of CGCM and SS-PAN (Fig.~\ref{fig:erf_p3}C), the PER increases from 0.008 to 0.090, an approximately $11\times$ increase over the P2/$-$P5 baseline. Relative to the unpruned YOLOv8s PER of 0.147, ScopeMamba-S recovers approximately 61\% of the measured peripheral ERF energy while retaining the high-resolution P2/$-$P5 scale allocation. This indicates that the proposed context pathway substantially extends the spatial reach of the high-resolution detection feature, consistent with its intended role in providing long-range contextual information for small-object detection.

\subsection{Ablation Study}
\label{sec:ablation}

\begin{figure}[htbp]
\centering
\includegraphics[width=\linewidth]{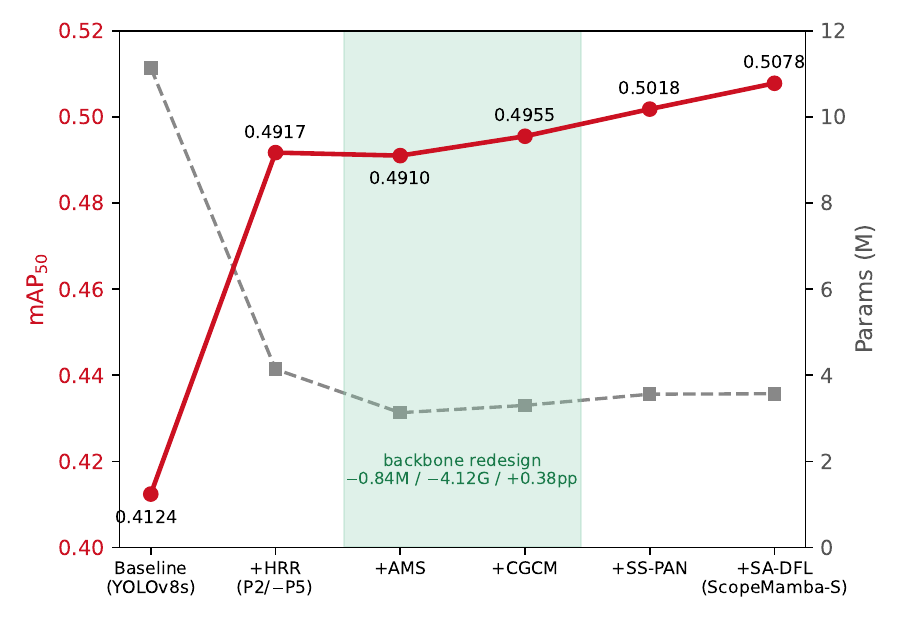}
\caption{Cumulative ablation trajectory on the VisDrone-2019 dataset (Small scale). The dual-axis design explicitly maps the correspondence between performance gains (left axis, $mAP_{50}$) and capacity reductions (right axis, Params in Millions) at each integration step.}
\label{fig:cumulative_ablation}
\end{figure}

To evaluate the contribution and resource cost of each component, we
conduct a cumulative ablation on VisDrone-2019, progressing from the
YOLOv8s baseline to ScopeMamba-S. Table~\ref{tab:cumulative_ablation}
reports mAP$_{50}$, parameter count, and GFLOPs at every step, while
Fig.~\ref{fig:cumulative_ablation} visualizes the corresponding
accuracy--parameter trajectory.

Across the full trajectory, the parameter count decreases by
approximately 68\% (from 11.14M to 3.57M) while mAP$_{50}$ increases
by 9.54 pp. The adopted HRR prior accounts for 7.93 pp of this gain by
moving the detection scales toward higher spatial resolution. Starting
from that reallocated baseline, the subsequent backbone, neck, and
regression redesign---including AMS, CGCM, SS-PAN, and SA-DFL---adds a further 1.61 pp while reducing the
parameter count by approximately 0.57M. The additional computational
cost introduced by the high-resolution pathway and subsequent modules is
reported directly in Table~\ref{tab:cumulative_ablation} and discussed
component by component below.

\begin{table*}[t]
\centering
\caption{Cumulative Ablation Experiment on the VisDrone-2019 Dataset (Small Scale). \\ $\Delta$ indicates the marginal change relative to the preceding step.}
\label{tab:cumulative_ablation}
\small
\begin{tabular}{l | cc | cc | cc}
\toprule
\textbf{Integration Step} & \textbf{Params (M)} & \textbf{$\Delta$ Params} & \textbf{GFLOPs} & \textbf{$\Delta$ GFLOPs} & \textbf{mAP$_{50}$} & \textbf{$\Delta$ mAP$_{50}$} \\
\midrule
1. Baseline (YOLOv8s)      & 11.14 & - & 28.45 & - & 0.4124 & - \\
2. + HRR (P2/$-$P5)        & 4.14  & $-$7.00 M & 51.10 & +22.65 G & 0.4917 & +0.0793 \\
3. + AMS-Block             & 3.13  & \textbf{$-$1.01 M} & 46.54 & \textbf{$-$4.56 G} & 0.4910 & $-$0.0007 \\
4. + CGCM                  & 3.30  & +0.17 M & 46.98 & +0.44 G & 0.4955 & +0.0045 \\
5. + SS-PAN                & 3.56 & +0.26 M & 49.43 & +2.45 G & 0.5018 & \textbf{+0.0063} \\
6. + SA-DFL (ScopeMamba-S) & 3.57 & +0.01 M & 53.69 & +4.25 G & \textbf{0.5078} & +0.0060 \\
\bottomrule

\end{tabular}
\end{table*}

\subsubsection{Effect of HRR (Reallocate)}
Replacing the stride-32 P5 detection level with a stride-4 P2 level
increases mAP$_{50}$ from 0.4124 to 0.4917, corresponding to a
7.93 pp gain, while reducing the parameter count from 11.14M to 4.14M.
The higher-resolution feature maps, however, increase computation from
28.45 to 51.10 GFLOPs. HRR therefore provides the main accuracy gain in
the cumulative trajectory, at the cost of substantially higher
computation.

\subsubsection{Effect of AMS-Block (Read)}
Adding the AMS-Block reduces the parameter count from 4.14M to 3.13M
and GFLOPs from 51.10 to 46.54, while mAP$_{50}$ changes only from
0.4917 to 0.4910 ($-0.07$ pp). To examine whether anisotropic kernel
geometry contributes beyond the parameter reduction, we replace the
strip kernels with isotropic square kernels of comparable receptive
extent in the final configuration. This control reaches 0.4963
mAP$_{50}$ compared with 0.5078 for the anisotropic configuration,
a difference of $-1.15$ pp at near-equal computational cost. The result
supports the use of directional strip kernels in the adopted AMS
configuration, while the associated parameter reduction helps offset
the cost of the subsequent context modules.

\subsubsection{Effect of SS-PAN (Reach)}

SS-PAN denotes the complete redesigned neck described in
Section~\ref{sec:method_neck}, comprising DySample-based top-down
fusion, directional P2 context injection, and bottom-up selective-scan
fusion. In the cumulative ablation, introducing SS-PAN increases
mAP$_{50}$ from 0.4955 to 0.5018, corresponding to a 0.63 pp gain for
an additional 0.26M parameters. To isolate the role of selective
scanning within the neck, we further remove only the bottom-up
selective-scan branches from the final configuration. This reduces
mAP$_{50}$ from 0.5078 to 0.4969, a 1.09 pp drop at the Small scale.

\begin{table}[htbp]
\centering
\caption{Scale-Dependent Contribution of the Selective-Scan Branches
within SS-PAN. A performance drop ($-\Delta$) when the scan branches
are removed indicates their contribution at that scale.}
\label{tab:sspan_cross_scale}
\setlength{\tabcolsep}{4pt}
\small
\begin{tabular}{lccc}
\toprule
\textbf{Scale} & \textbf{Scan ON} & \textbf{Scan OFF} & \textbf{$\Delta$ mAP$_{50}$} \\
\midrule
Nano   & 0.4394 & 0.4340 & $-$0.0054 \\
Small  & 0.5078 & 0.4969 & $-$0.0109 \\
Medium & 0.5263 & 0.5236 & $-$0.0027 \\
\bottomrule
\end{tabular}
\end{table}
  
To examine whether the contribution of the selective-scan branches
within SS-PAN is specific to the Small variant, we additionally remove
these branches from the Nano and Medium configurations. As shown in
Table~\ref{tab:sspan_cross_scale}, their removal reduces mAP$_{50}$ by
0.54 pp, 1.09 pp, and 0.27 pp for the N, S, and M variants,
respectively. The consistent direction of these changes indicates that
the selective-scan component of SS-PAN contributes positively across the
three evaluated scales, although the magnitude of the effect is
scale dependent.

\subsubsection{Effect of SA-DFL (Regress)}
\label{sec:bins}
\begin{figure}[htbp]
\centering
\includegraphics[width=\linewidth]{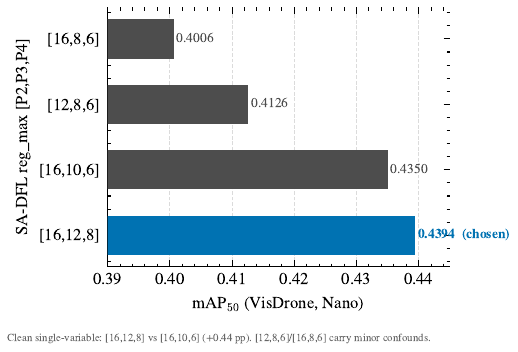}
\caption{Sensitivity analysis of the \texttt{reg\_max} bin allocation in the SA-DFL head evaluated at the Nano scale. Note that the x-axis starting point is set to 0.39 to distinctly highlight the fine-grained accuracy differences between structural configurations.}
\label{fig:regmax_sensitivity}
\end{figure}

The Scale-Adaptive Distribution Focal Loss (SA-DFL) head adds only
0.008M parameters and increases mAP$_{50}$ from 0.5018 to 0.5078,
corresponding to a 0.60 pp gain in the cumulative ablation. Its main
resource cost is computational rather than parametric. We further compare
different per-scale \texttt{reg\_max} allocations at the Nano scale in
Fig.~\ref{fig:regmax_sensitivity}; among the tested configurations, the
scale-adaptive allocation achieves the highest mAP$_{50}$. Since the
regression-branch width scales with $4K_i$
(Eq.~\ref{eq:sadfl_width}), changing the per-scale bin allocation also
reallocates regression capacity across feature levels. We therefore
interpret the observed gain as the effect of the joint scale-adaptive
regression design rather than as evidence for finer bin spacing alone.

Additional in-house regression-head and loss variants were also
evaluated during development, including alternative quality-branch
designs and NWD~\cite{xu2022detecting}. None provided a consistent
improvement over the adopted SA-DFL configuration in the tested
settings. These exploratory controls are therefore used as supporting
negative results rather than as evidence that the adopted regression
formulation is globally optimal.

\subsection{Leave-One-Out (LOO) Analysis and Robustness}
\label{sec:loo}

To characterize the run-to-run variation of the CGCM control, we train
the final configuration without CGCM using three random seeds. The three
runs obtain mAP$_{50}$ values of 0.5048, 0.5004, and 0.4980, yielding a
mean of 0.5011 and a standard deviation of 0.34 pp. The complete
ScopeMamba-S reaches 0.5078, which is 0.67 pp above this mean and also
above the highest value observed among the three control runs.

This multi-seed comparison and the cumulative ablation quantify
different effects. In Table~\ref{tab:cumulative_ablation}, adding CGCM
at Step~4 increases a single-run trajectory from 0.4910 to 0.4955,
corresponding to $+0.45$ pp. In the final configuration, the comparison
against the three-seed no-CGCM mean gives $+0.67$ pp. We therefore use
$+0.45$ pp when describing the cumulative integration trajectory and
$+0.67$ pp when discussing the final-configuration CGCM control.

Because multi-seed variance is measured only for the no-CGCM control,
the observed $\sigma=0.34$ pp is used as an empirical reference for
run-to-run variation rather than as a formal significance estimate for
all other architectural variants. Accordingly, comparisons such as
SS-PAN removal, in-path insertion, and isotropic-kernel substitution are
reported directly in percentage points rather than converted into
cross-variant $\sigma$ multiples.

\subsection{Discussion on Negative Results}
\label{sec:negative_results}
For completeness, we report several representative configurations that
did not improve the final model. These controls help delimit the design
choices supported by the experiments.

\textbf{Placement and Dimensional Constraints.}
Directly embedding selective scanning in the main backbone path reduces
mAP$_{50}$ by 0.98 pp in the matched early-stage control
(Section~\ref{sec:placement}). At the Nano scale, halving the scan width
in the neck reaches 0.4289 mAP$_{50}$, compared with 0.4326 when the
branch is removed entirely, while also using more parameters. Thus, the
tested tapered configuration provides no improvement in either accuracy
or parameter efficiency and does not offer a better trade-off than the
adopted configuration.

\textbf{Operator Substitution.}
At the backbone node, replacing the selective scan with a cost-matched
convolution produces 0.5015 mAP$_{50}$, close to the three-seed
no-branch mean of 0.5011, while EMA reaches 0.4979. At the neck node,
EMA reaches 0.5013 compared with 0.5078 for selective scanning. Among
the operators tested at these nodes, selective scanning therefore
provides the strongest and most consistent improvement, supporting its
use for long-range contextual modeling in the proposed off-path
formulation.

\textbf{Quality-Branch Stacking.}
Under the final flagship configuration with only the detection head
swapped, we evaluated in-house quality-branch designs based on
directional decomposition and cross-scale alignment, reaching 0.4987
and 0.4983, respectively---both approximately 0.9 pp below the SA-DFL
configuration (0.5078). Additional variants explored during development
(foreground-guided sparsity, edge-auxiliary supervision, and asymmetric
heads) were likewise not carried forward. These results suggest that,
under the tested configurations, the joint scale-adaptive regression design is more effective than adding score--quality alignment branches.

\textbf{Loss Shaping.}
Similarly, layering specialized localization losses (e.g., Inner-IoU
or degenerate-box variants) on top of the flagship SA-DFL
configuration reduces mAP$_{50}$ to 0.4980 and 0.4973, respectively,
both more than 0.9 pp below the flagship result of 0.5078. In a
separate matched pairing, NWD yields a marginal $+0.04$ pp change in
mAP$_{50}$ while reducing mAP$_{50-95}$ by 0.31 pp, indicating a
metric trade-off rather than a consistent localization improvement.
These results do not support adding the tested loss-shaping variants
to the final SA-DFL configuration.

\textbf{Capacity Reallocation Controls.} Finally, reallocating parameter budgets toward network depth rather than CGCM-based context injection did not improve performance ($-0.43$ pp at the Small scale), suggesting that additional depth does not provide the same benefit as the proposed context pathway. A related spatial control replaces the anisotropic strip kernels of AMS with isotropic square kernels of equivalent receptive field (Sec.~\ref{sec:ablation}). This substitution results in a 1.15 pp drop in accuracy, indicating that directional selectivity is a more effective design factor than raw kernel extent in this control.

\textbf{Summary.}
Taken together, these negative results favor the adopted off-path
selective-scanning formulation, scale-adaptive distributional regression,
and anisotropic feature extraction over the tested alternatives,
including direct main-path insertion, auxiliary quality branches,
additional loss shaping, and increased network depth.

\subsection{Qualitative Analysis}
\label{sec:qualitative}

\begin{figure*}[htbp]
\centering
\includegraphics[width=\textwidth]{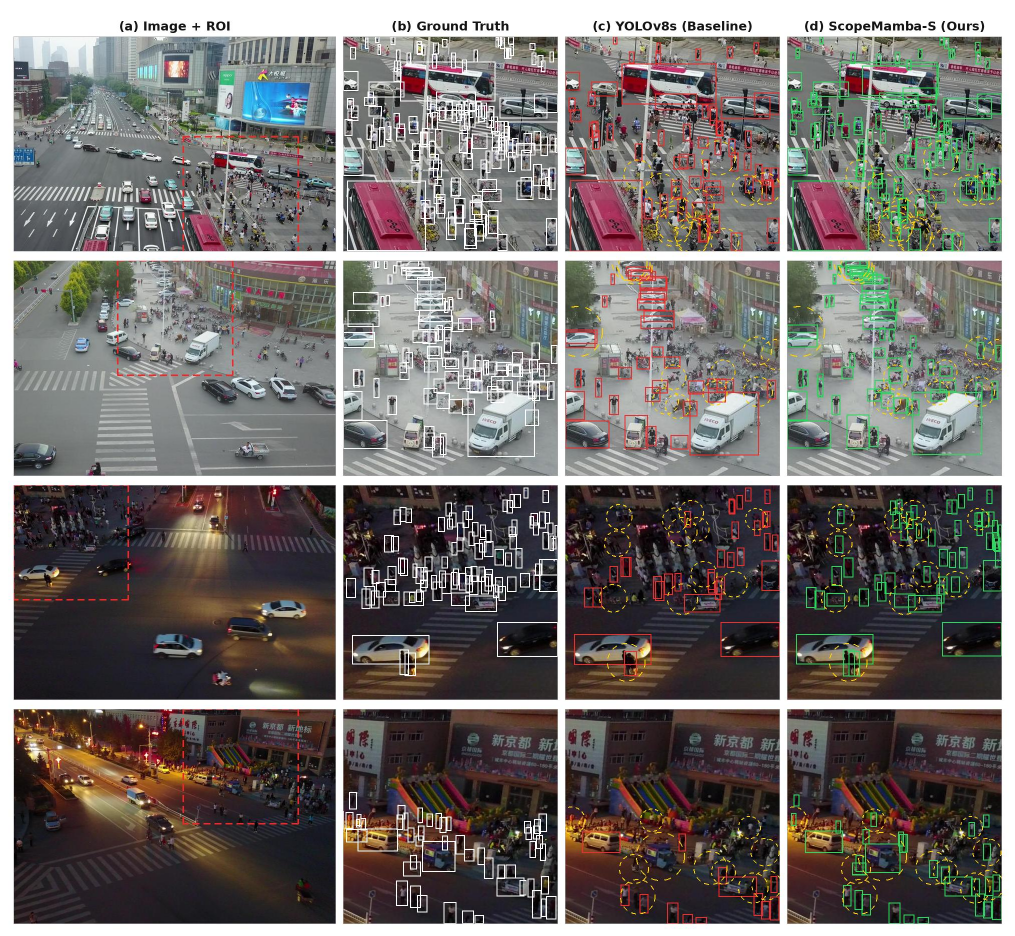}

\caption{Qualitative comparison on the VisDrone-2019 dataset.
(a) Original images with the region of interest (ROI) highlighted by a
red dashed box. (b) Ground-truth annotations. (c) Detection results of
YOLOv8s. (d) Detection results of ScopeMamba-S. Yellow dashed circles
mark representative annotated targets missed by YOLOv8s but detected
by ScopeMamba-S under dense, occluded, or low-illumination conditions.}

\label{fig:qualitative}
\end{figure*}

To qualitatively compare the two complete detectors,
Fig.~\ref{fig:qualitative} presents representative VisDrone-2019
validation scenes containing dense, occluded, and low-illumination
tiny objects. In the highlighted regions, ScopeMamba-S detects more
annotated pedestrians and vehicles than the YOLOv8s baseline,
including several examples missed under dense crowding, weak local
appearance cues, and low-light conditions. These complete-model observations are consistent with the quantitative
results; module-wise effects are evaluated separately through the
controlled experiments in Sections~\ref{sec:ablation}
and~\ref{sec:placement}.

\section{Conclusion}

This article investigates small-object detection in UAV and remote sensing imagery as a trade-off between preserving high-resolution object evidence and maintaining sufficiently broad contextual support. ScopeMamba-YOLO addresses this trade-off through off-path, zero-gated selective scanning, instantiated by CGCM in the backbone and selective-scan fusion in SS-PAN, while preserving the fine-grained convolutional stream. The AMS Block reduces the cost of high-resolution feature extraction, and SA-DFL reallocates distributional support and regression capacity across detection scales.

Controlled experiments support the proposed integration strategy: direct main-path selective scanning can be detrimental under a matched setting, whereas the off-path formulation provides positive gains that are not reproduced by cost-matched auxiliary branches. ERF analysis further shows that the proposed context pathway increases the peripheral energy ratio from 0.008 to 0.090 at the stride-8 feature, corresponding to an approximately $11\times$ expansion over the P2/$-$P5 baseline. Across VisDrone-2019 and AI-TOD, the four ScopeMamba variants achieve favorable accuracy--parameter trade-offs, with particularly clear improvements for very-tiny and tiny objects; ScopeMamba-M reaches the accuracy level of HEdge-MamYOLO with less than one-third of its parameters.

The remaining limitations mainly concern computation and generalization. The stride-4 pathway and native-resolution selective scanning in SS-PAN retain non-negligible computational cost, while the present evaluation is limited to $640\times640$ inputs without pretraining and to YOLO-style detection. Future work will therefore focus on reducing context-pathway latency and examining the proposed integration strategy under higher-resolution, pretrained, oriented, and multi-modal detection settings.



\ifCLASSOPTIONcaptionsoff
  \newpage
\fi



\bibliographystyle{IEEEtran}
\bibliography{ref}

\end{document}